\documentclass[twocolumn]{IEEEtran}

\usepackage{cite}
\usepackage{amsmath,amssymb,amsfonts}
\usepackage[normalem]{ulem}
\usepackage[thicklines]{cancel}
\usepackage{algorithmic}
\usepackage[ruled,vlined]{algorithm2e}
\usepackage{multicol}
\usepackage{dsfont}

\usepackage{graphicx}
\usepackage{amsthm}
\usepackage{subcaption}
\usepackage[flushleft]{threeparttable}
\usepackage[table]{xcolor}
\usepackage{multirow}
\usepackage{tikz}

\usepackage{comment}
\def\DefaultCutFileName{\def\CommentCutFile{\jobname.cut}}
\DefaultCutFileName

\newtheorem{theorem}{Theorem}

\newtheorem{lemma}[theorem]{Lemma}

\theoremstyle{definition}

\theoremstyle{remark}
\newtheorem{remark}{Remark}

\newcommand{\mc}[1]{\mathcal #1}

	\DeclareMathOperator*{\argmin}{arg\,min}
	\DeclareMathOperator{\E}{\mathbb{E}}
	
	\newcounter{relctr} 
	\everydisplay\expandafter{\the\everydisplay\setcounter{relctr}{0}} 

	\AtBeginDocument{} 
	\usepackage {xcolor}
	\newcommand{\yan}[1]{{\textcolor{black}{#1}}}

\begin{document}
		
		\title{Adversarial Combinatorial Bandits with  \\ Switching Costs}

\author{Yanyan Dong and Vincent Y. F. Tan, {\em Senior Member, IEEE}%
  \thanks{This work is supported by funding from a Ministry of Education Academic Research Fund (AcRF) Tier 2 grant under grant number A-8000423-00-00 and  AcRF Tier 1 grants under grant numbers  A-8000189-01-00 and A-8000980-00-00.}
  \thanks{Yanyan Dong is with the School of Science and Engineering, The Chinese University of Hong Kong, Shenzhen, Shenzhen 518172, China (email: yanyandong@link.cuhk.edu.cn). This work was carried out when Yanyan Dong was a Research Fellow at the Department of Electrical and Computer Engineering, 
  National University of Singapore.}
  \thanks{Vincent Y.~F.~Tan is with the  Department of Mathematics and the Department of Electrical and Computer Engineering, National University of Singapore (e-mail: vtan@nus.edu.sg), Singapore, 119077.}
    \thanks{The work was accepted in IEEE  Transactions on Information Theory.}
}
		
		\maketitle
	\begin{abstract}
		We study the problem of adversarial combinatorial bandit with a switching cost $\lambda$ for a switch of each selected arm in each round, considering both the bandit feedback and semi-bandit feedback settings.  In the oblivious adversarial case with $K$ base arms  and time horizon $T$, we derive lower bounds for the minimax regret and design algorithms to approach them. To prove these lower bounds, we design stochastic loss sequences for both feedback settings, building on an idea from previous work in Dekel et al. (2014). The lower bound for bandit feedback is $ \tilde{\Omega}\big( (\lambda K)^{\frac{1}{3}} (TI)^{\frac{2}{3}}\big)$    while that for semi-bandit feedback is $ \tilde{\Omega}\big( (\lambda K I)^{\frac{1}{3}} T^{\frac{2}{3}}\big)$     where $I$ is the number of base arms in the combinatorial arm played in each round. To approach these lower bounds, we design algorithms that operate in batches by dividing the time horizon into batches to restrict the number of switches between actions.
For the bandit feedback setting, where only the total loss of the combinatorial arm is observed, we introduce the \textsc{Batched-Exp2} algorithm which achieves a regret upper bound of $\tilde{O}\big((\lambda K)^{\frac{1}{3}}T^{\frac{2}{3}}I^{\frac{4}{3}}\big)$ as $T$ tends to infinity. In the semi-bandit feedback setting, where all losses for the combinatorial arm are observed, we propose the \textsc{Batched-BROAD} algorithm which achieves a  regret upper bound of $\tilde{O}\big( (\lambda K)^{\frac{1}{3}} (TI)^{\frac{2}{3}}\big)$.
		\end{abstract}

 \begin{IEEEkeywords}
     multi-armed bandits, adversarial bandits,  combinatorial bandits,  switching costs, online optimization
 \end{IEEEkeywords}
\section{Introduction}
The classical multi-armed bandit (MAB) problem is a sequential decision making game between an agent and an environment~\cite{lattimore2020bandit}, where the agent plays
the arms sequentially to  minimize the total loss over time.
 After each arm is played, the agent receives some feedback in the form of a loss (or gain) associated with the chosen arm. 
In many applications such as the financial trading or reconfiguration in industrial environments, there is a cost $\lambda>0$ for a switch of each selected arm in each round, which must be considered to assess the overall performance of the algorithms designed for them. For example, Guha and Munagala~\cite{guha2009multi} constructed a  sensor network problem and refined  probabilistic models of sensed
data at various nodes, which costs energy in  transferring from the current node
to a new node.  Shi et al.\ \cite{shi2022power} introduced an application of the switching cost in edge computing with artificial intelligence, where the edge server can only utilize a small number of machine learning models in each round to learn the best subset of models based on the feedback. Downloading a model that is not in the current edge server from the cloud incurs a switching cost.

The problem of MAB with switching costs has been studied for both the oblivious adversarial case and the stochastic case. We will focus on the oblivious adversarial setting, where losses are generated by an arbitrary deterministic source before the game.
In the oblivious adversarial case with $K$ arms and time horizon $T$, 
Arora et al.~\cite{arora2012online}  refined the Exp3 algorithm to achieve a regret upper bound\footnote{In this paper, we use $\tilde O$ and $\tilde \Omega$ to denote the big-O and big omega notations ignoring any logarithmic factors in $T$.} of $\tilde O(K^{\frac{1}{3}}T^{\frac{2}{3}})$ when the switching cost $\lambda=1.$ Later, Dekel et al.~\cite{dekel2014bandits} proved that the upper bound is tight by showing that the minimax regret lower bound is $\tilde{\Omega} ((\lambda K)^{\frac{1}{3}}T^{\frac{2}{3}})$. Rouyer et al.~\cite{rouyer2021algorithm} proposed an algorithm which is a modification of the Tsallis-Switch algorithm to achieve an upper bound of $O((\lambda K)^{\frac{1}{3}} T^{\frac{2}{3}})$. Without switching costs, the minimax regret of the adversarial MAB problem is $\Theta(\sqrt{TK})$~\cite{auer2002nonstochastic,cesa2006prediction}.

Combinatorial bandits form a general extension of the standard framework, which is a linear bandit with the special combinatorial action set $\mc A\subseteq \{0,1\}^K$.  We generalize the problem of MAB with switching costs by considering the combinatorial problem with $I$ arms  played in each round, where the set of the played arms is called a \emph{combinatorial arm}. 
\yan{There are many practical applications of combinatorial problems with switching costs. For example, a hospital may plan   to experiment on
		a drug that is known to be a combination of $I$ components  of  $K$ (raw material) components. In our parlance,  there are $K$ base arms and the combinatorial arm contains $I$ out of the $K$ base arms. 
		The quality of each chosen component depends on certain unknown complex effects of the environment and patients, and this may be modeled by an adversarial setting. The overall effect of the drug is the sum of the qualities of all the individual chosen components.
		There is, however, a non-negligible purchasing (or switching) cost when the agent decides to swap one component for another from one time step to the next.} In~\cite{shi2022power}, a special combinatorial problem was considered when the costly full feedback was available and a switching cost was added in each round. Our setting does not allow full feedback and only considers bandit feedback and semi-bandit feedback. Under bandit feedback, the player can only observe the total loss of the played combinatorial arm while under semi-bandit feedback, all the losses for the combinatorial arm  are observed.

In this paper, we will focus on analyzing the regret of adversarial combinatorial bandit when there are switching costs. 
  We  derive  lower bounds for the minimax regret and propose the algorithms that approximately meet the lower bounds under both feedback scenarios. 
To prove the lower bounds, we design  stochastic loss sequences   for both bandit feedback and semi-bandit feedback,  which generalize the idea in~\cite{dekel2014bandits} for the combinatorial scenarios.  Under  different types of feedback, 
the loss sequences designed are different. For a fixed time, the  loss sequence under bandit feedback uses the same Gaussian noise for different base arms while  the loss sequence under semi-bandit feedback uses i.i.d. Gaussian noises for different base arms. We show that the lower bound for bandit feedback is $ \tilde{\Omega}\big( (\lambda K)^{\frac{1}{3}} (TI)^{\frac{2}{3}}\big)$    while that for semi-bandit feedback is $ \tilde{\Omega}\big( (\lambda KI)^{\frac{1}{3}} T^{\frac{2}{3}}\big)$     where $I$ is the number of base arms in the combinatorial arm played in each round. 
Dekel et al.~\cite{arora2012online}, Rouyer et al.~\cite{rouyer2021algorithm} and  Shi et al.~\cite{shi2022power} all utilize the batch-based algorithms to restrict the number of switches between actions by dividing the whole time horizon into batches and forcing the algorithm to play the same action for all the rounds within a batch.
We also utilize this technique in our algorithms under our combinatorial setting. In the bandit feedback setting, we introduce the \textsc{Batched-Exp2} algorithm with John's exploration, which is a batched version of the Exp2 algorithm with John's exploration~\cite{bubeck2011introduction} and achieves a regret bound of $	 \tilde{O}\big( (\lambda K)^{\frac{1}{3}} T^{\frac{2}{3}}I^{\frac{4}{3}} \big)$ when the time horizon $T$ tends to infinity. In the semi-bandit feedback setting, we introduce the \textsc{Batched-BROAD} algorithm, which is a batched version of the Online Mirror Descent algorithm with log-barrier regularizer (BROAD) in~\cite{wei2018more} and achieves a regret upper bound of $	 \tilde{O}\big( (\lambda K)^{\frac{1}{3}}(TI)^{\frac{2}{3}}\big)$.

In the remainder of this paper, we first formulate the problem and introduce our main results in \S\ref{sec:formulationresults}. Then \S\ref{sec:lowerbound} presents the main ideas for proving the lower bound for two different types of feedback. 
In \S\ref{sec:banditsemi}, we introduce two algorithms designed for the bandit feedback and semi-bandit feedback respectively.  \S\ref{sec:numerical} is dedicated to the compare our algorithms with some baselines by numerical experiments. In Appendix, we provide complete proofs for the lower bounds.

\section{Problem Formulation  and Main Results}\label{sec:formulationresults}
We use $[n]$ to denote the set $\{1,\dots, n\}$ for any positive integer $n$ and $x_{1:r}$ to denote the sequence $\{x_i\}_{i=1}^r$.
We consider an adversarial combinatorial bandit problem with $K$ base arms $[K]$ and a switching cost $\lambda$ for each switched arm with $\lambda >0$. The loss vectors $l_t\in [0,1]^K$ for $t\geq 1$ are arbitrarily generated by the adversary and do not depend on the actions taken by the learner, where the $i$-th component of $l_t$ is the loss incurred if arm $i$ is pulled at time $t$.
Let $\mc A=\{A\in \{0,1\}^K: \lVert A \rVert_1 =I \}$ be the set of all combinatorial arms with $I$ base arms where the $i$-th components of $A$ is one if arm $i$ is pulled and zero otherwise. For the sake of simplicity, we define the set of base arms within a combinatorial arm $\mathcal{I} \in \mathcal{A}$ as $\{i \in [K]: \mathcal{I}_i = 1\}$, where $\mc I_i$ is the $i$-th component in $\mc I$.
 In each time $t\geq 1$, the player pulls a combinatorial arm $ A_t\in \mc A$ and incurs a loss  of $  \langle A_t,   l_t\rangle =\sum_{i=1}^K A_{t,i}l_{t,i}$ ($A_{t,i}$ and $l_{t,i}$ are respectively the $i$-th component of the vectors $A_t\in \{0,1\}^K$ and $ l_t\in [0,1]^K$), where $ \langle \cdot, \cdot \rangle $ denotes the inner product operation. We consider both bandit and semi-bandit feedback~\cite{combes2015combinatorial}. Under bandit feedback and after the combinatorial arm $A_t$ is pulled at  round $t$, the player can only observe the feedback $X_t= \langle A_t,   l_t\rangle \in [0,I]$ while under semi-bandit feedback, all the losses for the combinatorial arm $X_t=A_t \circ l_t\in [0,1]^K$ are  observed where $\circ$ stands for the element-wise multiplication.
From time $t-1$ to $t$,  there is a switching cost of $\lambda \cdot d(A_t, A_{t-1})$  for the player where $d(A_t, A_{t-1})\triangleq 
\frac{1}{2}\lVert A_t \oplus A_{t-1}\rVert_1$, i.e., $d(A_t, A_{t-1})$ measures the number of arms switched from the combinatorial arm $A_{t-1}$ pulled at time $t-1$ to $A_t$ pulled at time $t$. We set $A_0= 0$ and then the first action $A_1$ will always incur a switching cost of $\lambda I$. The cumulative loss over $T$ rounds for the player equals 
\begin{equation*}
\sum_{t=1}^T    \langle A_t,   l_t\rangle + \lambda \sum_{t=1}^T   d(A_t,A_{t-1}).
\end{equation*}
Given the loss sequence $l_{1:T}\in [0,1]^{K\times T}$ and action sequence $A_{1:T} \in \mc A^{T}$, define the $\lambda$-switching regret as
\begin{align*}
&	R_{\lambda}(A_{0:T},l_{1:T})\\
& \quad\triangleq \sum_{t=1}^T  \langle A_t, l_t\rangle+  \lambda \sum_{t=1}^T   d(A_t,A_{t-1}) - \min_{A\in \mc A} \sum_{t=1}^T   \langle A,l_t\rangle.
\end{align*}
A policy $\pi =\pi_{1:T}$ is composed of all the conditional distributions over actions at each time $t$ given past actions and feedback, i.e., $\pi_t (\cdot | A_1,X_1,\dots, A_{t-1},X_{t-1})$. The set of all policies in $\mc A^T$ is denoted as $ \Pi$ and the set of all deterministic loss sequences in $ [0,1]^{K\times T}$ is denoted as $ \mc L$.
We define the expected $\lambda$-switching regret  when the loss functions $l_{1:T}\in \mc L$ are specified and a policy $\pi\in \Pi$ is employed as follows,
\begin{align*}
	R_{\lambda}(\pi,l_{1:T}) = \E \big[R_{\lambda}(A_{0:T},l_{1:T})\big], \quad 
\end{align*}
where the expectation is taken over the player’s randomized choice of actions under the policy $\pi$. In this paper, we use regret to represent for $\lambda$-switching regret for simplicity. When $\lambda=0$, i.e., the switching cost is not considered, we write $R(\pi, l_{1:T}) = R_0(\pi, l_{1:T})$ for simplicity and we call $R(\pi, l_{1:T})$ the \emph{pseudo-regret}.
We use the \emph{minimax expected $\lambda$-switching  regret} to measure the difficulty of the game as in \cite{dekel2014bandits}, which is defined as follows,
\begin{align*}
	R_{\lambda}^{*} = \inf_{\pi \in \Pi} \sup_{l_{1:T}\in \mc L} 	R_{\lambda}(\pi, l_{1:T}).
\end{align*}

Under bandit feedback, the following theorem states that there  exists a loss sequence $l_{1:T}$ such that $R_{\lambda}(\pi,l_{1:T})$ is $\Omega\Big( \frac{(\lambda K)^{\frac{1}{3}} (TI)^{\frac{2}{3}}}{\log_2 T}\Big)$ for any player policy $\pi\in \Pi$, which implies that $R_{\lambda}^*=\Omega\Big( \frac{(\lambda K)^{\frac{1}{3}} (TI)^{\frac{2}{3}}}{\log_2 T}\Big)$.
\begin{theorem}\label{thm:lowerbound}
  	Consider the combinatorial bandit with switching costs under the bandit feedback. For any  player policy $\pi\in \Pi$, there exists a loss sequence $l_{1:T}\in \mc L$  that incurs an expected $\lambda$-switching regret of 
	$$R_{\lambda}(\pi,l_{1:T})= \Omega\Bigg( \frac{(\lambda K)^{\frac{1}{3}} (TI)^{\frac{2}{3}}}{\log_2 T}\Bigg), $$
	provided that $K\geq 3I$ and  $ T\geq \max \{\frac{\lambda K}{I},8\} $.
\end{theorem}

Under semi-bandit feedback,
the following theorem states that there always exists a loss sequence $l_{1:T}$ such that $R_{\lambda}(\pi,l_{1:T})$ is $\Omega( \frac{(\lambda KI)^{\frac{1}{3}} T^{\frac{2}{3}}}{\log_2 T})$ for any player policy $\pi\in \Pi$, which implies that $R_{\lambda}^*=\Omega( \frac{(\lambda KI)^{\frac{1}{3}} T^{\frac{2}{3}}}{\log_2 T})$.
\begin{theorem}\label{thm:lowerbound-semi}
	  	Consider the combinatorial bandit with switching costs under the semi-bandit feedback. For any  player policy $\pi\in \Pi$, there exists a loss sequence $l_{1:T}\in \mc L$ that incurs an expected $\lambda$-switching  regret of 
	$$R_{\lambda}(\pi,l_{1:T})= \Omega\Bigg( \frac{(\lambda K I)^{\frac{1}{3}} T^{\frac{2}{3}}}{\log_2 T}\Bigg), $$
	provided that $K\geq 3I$ and $ T\geq  \max \{\frac{\lambda K}{I^2},6\} $.
\end{theorem}
We  observe that the orders of the lower bound of regret in Theorems~\ref{thm:lowerbound} and~\ref{thm:lowerbound-semi} differ only in the combinatorial size $I.$
Also, the lower bound in Theorem~\ref{thm:lowerbound} is greater than that in Theorem~\ref{thm:lowerbound-semi} in terms of the order of  $I$ due to the fact that semi-bandit feedback results in more observations for the player compared with bandit feedback.

To prove Theorems~\ref{thm:lowerbound} and~\ref{thm:lowerbound-semi}, we design two stochastic loss sequences for bandit feedback and semi-bandit feedback respectively, which are presented in Algorithms~\ref{alg:sequence1} and~\ref{alg:sequence2} of Sections~\ref{subsec:lowerboundbandit} and~\ref{subsec:lowerbound-semi}, respectively. The design of two loss sequences generalizes the idea in~\cite{dekel2014bandits} for the combinatorial scenarios. For both loss sequences, we  apply Yao's minimax principle~\cite{yao1977probabilistic}, which asserts that the regret of a randomized player against the worst-case adversary is at least the minimax regret of the optimal deterministic player against a stochastic loss sequence.  By employing this principle, we are able to  prove the two theorems using the constructed adversaries. Under  different types of feedback, 
the adversaries designed are different. For a fixed time, the  adversary under bandit feedback uses the same Gaussian noise for different base arms while  the adversary under semi-bandit feedback uses i.i.d. Gaussian noises for different base arms.

We also design two algorithms  for two types of feedback that can be shown 
 to be almost optimal when compared to the lower bounds. Similar to~\cite{arora2012online,rouyer2021algorithm,shi2022power}, we utilize the batched algorithm to limit the number of switches between actions by dividing the whole time horizon $T$ into batches and forcing the algorithm to play the same combinatorial arm for all the rounds within a batch.
For bandit feedback, the \textsc{Exp2} with John's exploration algorithm~\cite{bubeck2011introduction} is the most efficient among the existing algorithms when switching cost is not considered.
We introduce a refined version of this algorithm called the \textsc{Batched-Exp2} algorithm with John's exploration when the switching cost is involved, which achieves a regret bound  as shown in below.

\begin{theorem}[Informal]\label{thm:exp2informal}
    	Consider the combinatorial bandit problem under bandit feedback.  For any adversary $l_{1:T}\in \mc L$, the policy $\pi$ of \textsc{Batched-Exp2} with John's exploration distribution as detailed in Algorithm~\ref{alg:EXP2John} achieves a $\lambda$-switching regret of
\begin{equation*}
    R_{\lambda}(\pi,l_{1:T}) =O  \big((\lambda K)^{\frac{1}{3}} T^{\frac{2}{3}}I^{\frac{4}{3}} \big).
\end{equation*}
\end{theorem}

For semi-bandit feedback, the BROAD algorithm~\cite{wei2018more} achieves the minimax regret when switching cost is not considered.
We introduce a refined version of this algorithm called the \textsc{Batched-BROAD} algorithm when the switching costs are involved, which  achieves a regret bound  as shown in below. 

\begin{theorem}[Informal]\label{thm:broadinformal}
Consider the combinatorial bandit problem under semi-bandit feedback.    For any adversary $l_{1:T}\in \mc L$, the policy $\pi$ of \textsc{Batched-BROAD}  as detailed in Algorithm~\ref{alg:BROAD} achieves a $\lambda$-switching regret of
    \begin{equation*}
        R_{\lambda}(\pi,l_{1:T})=\tilde O\big( (\lambda K)^{\frac{1}{3}}(TI)^{\frac{2}{3}}+KI\big).
    \end{equation*}
\end{theorem}

We compare the lower bounds and upper bounds in Table~\ref{table:comparison}, which shows that the regret gap between \textsc{Batched-Exp2} with John's exploration and the lower bound scales at most as $I^{\frac{2}{3}}$ and that between \textsc{Batched-BROAD} and the lower bound scales at most $I^{\frac{1}{3}}$. Closing these gaps appears to be challenging and is left for future work.

\begin{table}[t]
\centering
\caption{Comparison Between the Lower Bounds and Upper Bounds under Two Types of Feedback}\label{table:comparison}
\resizebox{\columnwidth}{!}{%
\begin{tabular}{ccc}
\hline
            & bandit feedback & semi-bandit feedback \\ \hline
lower bound & $ \tilde{\Omega}\big( (\lambda K)^{\frac{1}{3}} (TI)^{\frac{2}{3}}\big)$                               & $\tilde\Omega\big( (\lambda K I)^{\frac{1}{3}} T^{\frac{2}{3}}\big)$                    \\ \hline
upper bound & $ O\big( (\lambda K )^{\frac{1}{3}} T^{\frac{2}{3}}I^{\frac{4}{3}}\big)$               & $\tilde O\big( (\lambda K )^{\frac{1}{3}} T^{\frac{2}{3}}I^{\frac{2}{3}}+KI\big)$                    \\ \hline
\end{tabular}}
\end{table}

\section{Lower Bound Analysis}\label{sec:lowerbound}

Following the method in \cite{dekel2014bandits}, we  apply Yao's minimax principle~\cite{yao1977probabilistic} to prove Theorems~\ref{thm:lowerbound} and~\ref{thm:lowerbound-semi}. The principle states that the regret of a randomized player against the worst-case loss sequence is at least the minimax regret of the optimal deterministic player against a stochastic loss sequence. Thus Theorem~\ref{thm:lowerbound} (resp. Theorem~\ref{thm:lowerbound-semi}) holds if we can construct a stochastic sequence of  loss vectors  $L_{1:T}$ (each $L_t = \big(L_{t,1},\dots, L_{t,K}\big)\in [0,1]^K$ is a random vector) such that 
\begin{align}
&	R_{\lambda}(\pi, L_{1:T})\nonumber\\*
 &\triangleq	\E \Bigg [\sum_{t=1}^T  \langle A_t, L_t\rangle + \lambda  \sum_{t=1}^T   d(A_t,A_{t-1}) \Bigg]- \min_{A\in \mc A} \sum_{t=1}^T \langle A, L_t\rangle \label{eq:defRlambdasto}\\
	&=\Omega\Big( \frac{(\lambda K)^{\frac{1}{3}} (TI)^{\frac{2}{3}}}{\log_2 T}\Big) \quad \Bigg(\text{resp. } \Omega\Big( \frac{(\lambda K I)^{\frac{1}{3}} T^{\frac{2}{3}}}{\log_2 T}\Big)\Bigg),\nonumber
\end{align}
for any deterministic player policy $\pi$, where the expectation is taken over the adversary’s randomized choice of loss vectors.
 In the following two subsections, we will judiciously construct specific loss sequences for the two feedback scenarios, which are key to proving Theorems~\ref{thm:lowerbound} and~\ref{thm:lowerbound-semi}.
 
\subsection{Proof of Theorem~\ref{thm:lowerbound}: Bandit Feedback}\label{subsec:lowerboundbandit}
		  \begin{algorithm}[t]
	\SetKwInOut{Input}{Input}
	\SetKwInOut{Output}{Output}
	\SetAlgoLined
	\Input{Time horizon $T$, switching cost $\lambda$, number of base arms $K$ and combinatorial arm size $I$.}  
	\begin{enumerate}
		\item[\emph{Step 1:}] Set \\
   \begin{minipage}{0.42\textwidth}
  \begin{align}
  \epsilon &= \frac{(\lambda K)^{\frac{1}{3}}(IT)^{-\frac{1}{3}}}{9\log_2 T}, \label{eq:sigmaeps1}\\
  \sigma &=\frac{1}{6 \sqrt{ \log_2 T \log_2 \frac{4T(\lambda+\epsilon)}{\epsilon}}}  .\nonumber
  \end{align}
  Choose $ \chi \in \mc A$ uniformly at random and then
 generate $W_t$, for $t=[T]$ according to~\eqref{eq:wdef}.  
 \end{minipage}
		\item[\emph{Step 2:}]
  For all $ t\in [T] $ and $x\in [K]$, set  $x$-th components of 
		  $\tilde L_t= \big(\tilde L_{t,1},\dots, \tilde L_{t,K}\big)$ and $L_t= \big(L_{t,1},\dots, L_{t,K}\big)$
    as
    \begin{minipage}{0.42\textwidth}
             \begin{align}\label{eq:Lprime}
			\tilde{L}_{t,x} = W_t +\frac{1}{2} - \epsilon \chi_x, \quad 
			L_{t,x}  = \text{clip} (\tilde{L}_{t,x}).
		\end{align}
        \end{minipage}
	\end{enumerate}
	\Output{Loss sequence $L_{1:T}$.}
	
	\caption{The Combinatorial Identical-Noise (CIN) loss sequence}\label{alg:sequence1}
\end{algorithm}

   \begin{figure}[!t]
\centering
\resizebox{\columnwidth}{!}{%
\begin{tikzpicture}
\path (0.5, 0) coordinate (A) node[below] {$0$};
\path (2, 0) coordinate (B) node[below] {$1$};
\path (3.5, 0) coordinate (C) node[below] {$2$};
\path (5, 0) coordinate (D) node[below] {$3$};
\path (6.5, 0) coordinate (E) node[below] {$4$};
\path (8, 0) coordinate (F) node[below] {$5$};
\path (9.5, 0) coordinate (G) node[below] {$6$};
\path (10, 0) coordinate (H) node[below] {$t$};
\draw (0,0) edge[->] (10,0);
    \path[->] (A) edge [bend left] node [right] {} (B);
     \path[->] (A) edge [bend left] node [right] {} (C);
     \path[->] (C) edge [bend left] node [right] {} (D);
     \path[->] (A) edge [bend left] node [right] {} (E);
     \path[->] (E) edge [bend left] node [right] {} (F);
     \path[->] (E) edge [bend left] node [right] {} (G);
\end{tikzpicture}}
\caption{An illustration of the definition of parent time  $\rho(t)$. There is an arrow from each time $t$-th parent time $\rho(t)$ to $t$. For example,   $\rho(3)=2$ and $\rho(4)=0$.}\label{fig:rho_def}
\end{figure}
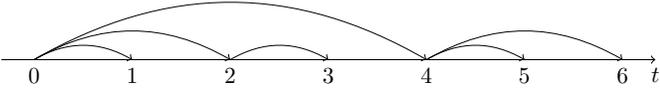

In this section, we provide the proof of Theorem~\ref{thm:lowerbound}.
First we obtain Lemma~\ref{lm:lowerbound1} for
  the  stochastic sequence of loss vectors $L_{1:T}$ in Algorithm~\ref{alg:sequence1} which is constructed by generalizing the loss sequence in~\cite{dekel2014bandits}.  Let clip$ (\alpha ) := \min \{ \max \{\alpha,0\},1\} $ and let $\chi_x$ denote the $x$-th coordinate of a vector $\chi\in \mc A$. Let the \emph{parent time} of $t$ be
  \begin{align}\label{eq:rhodef}
  \rho(t) = t- 2^{\delta(t)} ,\quad \delta(t) = \max \{i
  \geq 0: 2^i \text{ divides } t \}.
  \end{align}
 See Figure~\ref{fig:rho_def} for an illustration of $\rho(t)$.
 We say a time slot $t'$ is an \emph{ancestor} for $t$ if $t'= \rho^{(c)}(t)$ for some positive integer $c$ and $\rho^{(c)}$ stands for the $c$ iterated composition of $\rho.$ 
Given the function $\rho$, recursively define as in~\cite{dekel2014bandits} \emph{the set of all ancestors} of $t$ as $\mc S(t)$ by $\mc S(0)=\emptyset$ and 
\begin{align}\label{eq:rhostar}
	\mc S(t) = \mc S(\rho(t))\cup \{\rho(t)\}, \quad t\in [T].
\end{align}
The \emph{depth} of $\rho$ is then defined as $d(\rho)=\max_{t\in [T]} |\mc S(t)|$.
As in~\cite{dekel2014bandits}, we define $
    \mathrm{cut}(t)=\{s\in [T]:\rho(s)<t\leq s\},$
which is the set of time slots that are separated from their parent by $t.$ The \emph{width} of $\rho$ is defined as
\begin{equation}\label{eq:width}
    w(\rho) = \max_{t\in [T]} |\mathrm{cut}(t)|.
\end{equation}
\yan{The design of $\rho$ in~\eqref{eq:rhodef} guarantees that the depth $d(\rho)$ and width $w(\rho)$ are both upper bounded by $\log_2 T +1$~\cite{dekel2014bandits}.}
 Define $ W_{t}$ for $ t=1,\dots ,T $ recursively by setting $  	W_0= 0$ and
  \begin{align}\label{eq:wdef}
  	W_t &= W_{\rho(t)}+\xi_t,\quad \forall\,  t\in [T],
  \end{align}
  where $ \xi_t$, $ t=1,\dots,T $ are independent zero-mean, variance $ \sigma^2 $ Gaussian veriables.  \yan{The design of the parent time function $\rho(t)$ and $W_t$ guarantees that $\tilde{L}_{t,x}$ lies in $ [0,1]$ with high probability,  which allows us to control
   the difference between the $\lambda$-switching regrets under $L_{1:T}$ and $\tilde L_{1:T}$ 
   when the same deterministic strategy is applied. Then we can first analyze the regret bound under the loss sequence $\tilde L_{1:T}$, which is easier due to the fact that the random variables in it are sums of Gaussian random variables. This allows us to bound the regret under $L_{1:T}$.} The detailed proof for this guarantee is provided in Lemma~\ref{lm:RRprime_v2} of Appendix~\ref{app:banditlower}.

  The differences between the CIN loss sequence in Algorithm~\ref{alg:sequence1} and the loss sequence  in~\cite{dekel2014bandits} include the presence of $I$ best arms in each round instead of one and the variations in the values of the parameters $\epsilon$ and $\sigma$.	As in~\cite{dekel2014bandits}, the definition of $W_t$ induces the common uncertainty of all base arms. At each time $t$, the losses of the base arms in $\chi$ are all the same and greater than other base arms by a constant $\epsilon.$ As the player can only observe the total loss of all the base arms in the chosen combinatorial arm, it is difficult to figure out whether the loss observed is induced by the randomness of $W_t$ or due to the chosen arms being better than other ones if the algorithm does not switch the chosen combinatorial arm for some time. Therefore, by the constructed loss sequence $L_t$, we can prove the lower bound in Theorem~\ref{thm:lowerbound}.
  
  By Yao's minimax principle, Theorem~\ref{thm:lowerbound} can be proved if the following lemma is verified.
  \begin{lemma}\label{lm:lowerbound1}
  	Consider the combinatorial bandit problem with switching costs under the bandit feedback.
  	Let $L_{1:T}$ be the stochastic sequence of loss vectors defined in Algorithm~\ref{alg:sequence1}.
  	When $K\geq 3I$ and 
   $ T\geq \max \{\frac{\lambda K}{I},8\} $, for any deterministic player's policy $\pi$, we have 
  	$$	R_{\lambda}(\pi, L_{1:T})\geq 
   \frac{(\lambda K)^{\frac{1}{3}} (TI)^{\frac{2}{3}}}{260\log_2 T}. $$
  \end{lemma}
  

    
       
       \yan{To prove Lemma~\ref{lm:lowerbound1},    we need to analyze the expected regret under an arbitrary deterministic policy $\pi: \mathbb R^{I\times T}\to \mc A^T$ when the loss sequence is $L_{1:T}$. Note that under $ L_{1:T}$,  a deterministic policy $\pi$ yields an action sequence $ A_{1:T} \in \mc A^T$ so that $A_t$ is a function of the player's past observations $X_{1:t-1} $ with $X_t  = \langle A_t, L_t\rangle $.  We first analyze the expected regret under the same deterministic policy $\pi$ and the loss sequence $\tilde{L}_{1:T}$, which is defined 
       in~\eqref{eq:Lprime} of Algorithm~\ref{alg:sequence1}. Similar to $L_{1:T}$, the deterministic policy $\pi$ yields an action sequence $\tilde A_{1:T} \in \mc A^T$ so that $\tilde A_t$ is a function of past observations $Y_{0:t-1}$ under $\tilde L_{1:T}$,  where }
\begin{equation}\label{eq:ytsequence_v2}
       \yan{Y_0=1/2, \quad Y_t= \langle \tilde A_t, \tilde{L}_t\rangle.}
       \end{equation}
	Define the conditional probability measures
\begin{align}
	\mathcal Q_{\mathcal I}(\cdot ) &= P(\cdot | \chi = \mathcal I), \quad \quad \mathcal I \in \mc A ,\label{eq:Qdef1}\\
 \mathcal Q_{0}(\cdot )& = P(\cdot | \chi = \emptyset), \label{eq:Qdef2}
\end{align}
where $\mc Q_{\mc I}$ and $\mc Q_0$ are the probability distributions under the adversaries with $\chi = \mc I$ and   $ \chi =\emptyset $, respectively. Thus $ \mathcal Q_0 (\cdot)$ is the probability distribution when all arms incur the same loss.  Let $ \tilde{\mathcal F} $
		be the $ \sigma $-algebra generated by the player's observations  $Y_{1:T} $. 
		The total variation distance between $ \mathcal Q_0 $ and $ \mathcal Q_{\mathcal I} $ over $\tilde{\mc F}$ is defined as		\begin{align}\label{eq:defdev_v3}
			d^{\tilde{\mathcal F}}_{\mathrm{TV}} (\mathcal Q_0, \mathcal Q_{\mathcal I})  = \sup_{A\in \tilde{\mathcal F}} | \mathcal Q_0 (A) - \mathcal Q_{\mathcal I}(A)  |.
		\end{align}
  This distance captures the player’s ability to identify whether combinatorial arm $\mc I$ is better than or equivalent to the other combinatorial
  arms based on the loss values he observes~\cite{dekel2014bandits}. In the following, we first give a key lemma that relates the player’s ability to identify the best action to the number of switches he performs to or from base arms in $ \mathcal I $.  
  
  Define $ Y_S= \{Y_t\}_{t\in S} $ and let $ \Delta (Y_S| Y_{S'}) $ be the KL divergence  between the distribution of $ Y_S $ conditioned on $ Y_{S'} $ under $ \mathcal Q_0$ and $ \mathcal Q_{\mathcal I} $, i.e.,
\begin{align}\label{eq:defKL}
	\Delta(Y_S| Y_{S'}) \triangleq \E_{\mathcal Q_0} \Bigg[ \ln \frac{\mathcal Q_0 (Y_S| Y_{S'})}{\mathcal Q_{\mathcal I}( Y_S| Y_{S'})} \Bigg].
\end{align}
Using the chain rule,
\begin{align*}
	\Delta(Y_{0:T}) \triangleq	\Delta(Y_{0:T} | \emptyset) 
	&=\sum_{t=1}^T   \Delta (Y_t |Y_{\mc S(t)}).
\end{align*}
The analysis in~\cite{dekel2014bandits} focused on the evaluation of $\Delta(Y_{0:T})$, i.e. the Kullback-Leibler (KL) divergence between  $\mc Q_0(Y_{0:T})$ and $\mc Q_{\mc I}(Y_{0:T})$, $d_{\mathrm{KL}}(\mc Q_0(Y_{0:T})\rVert \mc Q_{\mc I}(Y_{0:T}))$, which has a close connection to the number of switches of arms in their setting and therefore leads to the lower bound of the regret.  As~\cite{dekel2014bandits}, we define 
\begin{align}\label{eq:defMI}
M_{\mathcal I} \triangleq 2\sum_{t=1}^T  \langle A_t \oplus A_{t-1} ,\mc I\rangle \text{ and } \tilde M_{\mathcal I} \triangleq 2\sum_{t=1}^T  \langle \tilde A_t \oplus \tilde A_{t-1} ,\mc I\rangle, 
\end{align}
as the total numbers of switches the player performs to or from base arms in $ \mathcal I $ during the whole time horizon under $L_{1:T}$ and $\tilde L_{1:T}$, respectively. 
Define the total numbers of switches in the whole time horizon under $L_{1:T}$ and $\tilde L_{1:T}$ as 
\begin{align}\label{eq:defM}
    M\triangleq \sum_{t=1}^T d(A_t, A_{t-1}) \text{ and } \tilde M\triangleq \sum_{t=1}^T d(\tilde A_t, \tilde A_{t-1})
    .
\end{align}
\yan{Notice that for $\forall\, i\in [K]$, we have $\big|\{\mc I\in \mc A: \mc I_i=1\}\big| = \binom{K-1}{I-1}$, where $\mc I_i$ is the $i$-th component of $\mc I$, and then the sum over all $\mc I\in \mc A$  of $M_{\mc I}$ (i.e., $\sum_{\mc I\in \mc A} M_{\mc I}$) is $\binom{K-1}{I-1}$ times the total number of switches the player performs to or from base arms in $[K]$. Since each switch is counted twice in the ``to" and ``from" directions, respectively, we conclude  that $\sum_{\mc I\in \mc A} M_\mc I = 2\binom{K-1}{I-1}M$ and $\sum_{\mc I\in \mc A} \tilde M_\mc I = 2\binom{K-1}{I-1}\tilde M$.}

The upper bound for $\Delta(Y_{0:T})$ in terms of $\tilde M_{\mc I}$ and $w(\rho)$ can be derived as follows, which directly leads to the upper bound for $ d^{\tilde{\mathcal F}}_{\mathrm{TV}} (\mathcal Q_0, \mathcal Q_{\mathcal I})$ in terms of $\tilde M_{\mc I}$ and $w(\rho)$.  Based on Lemma~\ref{lm:partswitch1}, we can prove Lemma~\ref{lm:lowerbound1}, and the whole proof is provided in Appendix~\ref{app:banditlower}.
		\begin{lemma}\label{lm:partswitch1}
  Under the loss sequence $\tilde{L}_{1:T} $, which is defined in~\eqref{eq:Lprime} of Algorithm~\ref{alg:sequence1} with $\chi = \mc I\in \mc A$, it holds that 
  \yan{$ \Delta(Y_{0:T})   \leq \frac{\epsilon^2}{2I\sigma^2} w(\rho) \E_{\mathcal Q_0}[\tilde M_{\mathcal I}]$, which implies that $$d^{\tilde{\mathcal F}}_{\mathrm{TV}} (\mathcal Q_0, \mathcal Q_{\mathcal I})\leq \frac{\epsilon}{2\sigma\sqrt{I}} \sqrt{w(\rho)  \E_{\mathcal Q_0}[\tilde M_{\mathcal I}]}.$$ }
		\end{lemma}
  
\begin{IEEEproof}
For any combinatorial arm $A\in \mc A$,  $\langle A, \mc I\rangle$ is the number of optimal arms in the combinatorial arm.
Under $\mc Q_{\mc I}$, by~\eqref{eq:Lprime} and~\eqref{eq:wdef}, we have 
\begin{align*}
Y_{\rho(t)}=\langle \tilde A_{\rho(t)}, \tilde L_{\rho(t)}\rangle=
(W_{\rho(t)}+\frac{1}{2})I-\epsilon\cdot \langle \tilde A_{\rho(t)}, \mathcal I\rangle,
\end{align*}
and 
\begin{align*}
Y_{t}=\langle \tilde A_{t}, \tilde L_{t}\rangle=(W_{\rho(t)}+\frac{1}{2}+\xi_t)I-\epsilon\cdot \langle \tilde A_{t}, \mathcal I\rangle.
\end{align*}
Under $\mc Q_0$, similarly we have \begin{align*}
Y_{\rho(t)}=
\Big(W_{\rho(t)}+\frac{1}{2}\Big)I,\quad 
Y_{t}=\Big(W_{\rho(t)}+\frac{1}{2}+\xi_t\Big)I.
\end{align*}
  Then the distribution of $ Y_t $ conditioned on $ Y_{\mc S(t)} $ is $ N(Y_{\rho(t)}, I^2\sigma^2) $ under  $ \mathcal Q_0 $ and  $ N(Y_{\rho(t)} + N_t \epsilon, I^2\sigma^2) $ under $ \mathcal Q_{\mathcal I} $, where  $N_t=\langle \tilde A_{\rho(t)}, \mathcal I\rangle - \langle \tilde A_{t}, \mathcal I \rangle $ is the difference between the numbers of arms in $\mc I$ played at time $\rho(t)$ and $t$, and $\epsilon$ is defined in~\eqref{eq:sigmaeps1} of Algorithm~\ref{alg:sequence1}. 
			Therefore,
\begin{align*}
	\Delta (Y_t |Y_{\rho(t)}) &= \sum_{i=-I}^I \mathcal Q_0 (N_t =i) d_{\mathrm{KL}}\Big (N(0,I^2\sigma^2)\Big\rVert N(i\epsilon, I^2\sigma^2) \Big) \\
	&= \sum_{i=1}^I 	Q_0 (|N_t| =i)   \frac{i^2\epsilon^2}{2I^2\sigma^2},
\end{align*}
where $\{ |N_t| =i \}$ is the event that the player switched at least $ i $ arms from or to base arms in $ \mc I $ between rounds $ \rho (t) $ and $ t $. We observe that $ N_t'\triangleq\langle \tilde A_{\rho(t)}\oplus \tilde A_{t}, \mathcal I\rangle $ is the total number of switches the player performs to or from base arms in $ \mc I $ between rounds $ \rho(t) $ and $ t $. Then $|N_t|\leq N_t'$ and
\begin{align}
	\Delta(Y_{0:T}) &= \frac{\epsilon^2}{2I^2\sigma^2} \sum_{t=1}^T    \sum_{i=1}^I 	Q_0 (|N_t| =i)  i^2 \nonumber\\
	&\leq  \frac{\epsilon^2}{2I\sigma^2} \sum_{t=1}^T    \sum_{i=1}^I 	Q_0 (|N_t| =i)i\nonumber\\
	&= \frac{\epsilon^2}{2I\sigma^2} \sum_{t=1}^T   \E_{Q_0} [|N_t|]\nonumber\\
	&\leq \frac{\epsilon^2}{2I\sigma^2} \sum_{t=1}^T   \E_{Q_0} [N_t'].\label{ineq:ntandntprime}
\end{align}
 We have $N_t' = \sum_{i\in [K]} \mc I_i\cdot \mathds{1}_{Z_{t,i}}$, where  $Z_{t,i}=\{\tilde A_{\rho(t),i}\neq \tilde A_{t,i}\}$ ($\tilde A_{x,i}$ is used to denote the $i$-th component of $\tilde A_x$). Let $$M_i=\big|\{t\in [T]: \tilde A_{t-1, i}\neq \tilde A_{t,i} \text{ or }\tilde A_{t, i}\neq \tilde A_{t+1,i} \}\big|,$$  denote the total number of switches the player performs to or from action $i$ during the whole time horizon. We have $\tilde M_{\mc I}=\sum_{i\in [K]}\mc I_i \cdot M_i$.
Let $s_{1:M_i,i}$ denote the time slots of switches from or to arm $i$, i.e. $\tilde A_{s_{j,i}-1,i}\neq \tilde A_{s_{j,i}, i}$ or $\tilde A_{s_{j,i},i}\neq \tilde A_{s_{j,i}+1, i}$ for any $j\in \{1,\dots, M_i\}$. Since the event $Z_{t,i}$ implies that there exists at least one time $s$ of switch from or to action $i$, such that $t\in \mathrm{cut}(s)$, we have 
\begin{align}
    &\sum_{t=1}^T N_t'=\sum_{t=1}^T \sum_{i\in [K]} \mc I_i\mathds{1}_{Z_{t,i}} \leq  \sum_{i\in [K]} \mc I_i  \sum_{t\in \mathrm{cut}(s_{r,i})} \mathds{1}_{Z_{t,i}} \nonumber\\*
    &\leq     \sum_{i\in [K]} \mc I_i \sum_{r=1}^{M_i} |\mathrm{cut}(s_{r,i})|\leq  \sum_{i\in [K]} \mc I_i M_i w(\rho) = \tilde M_{\mc I}w(\rho),\label{eq:banditswitch}
\end{align}
where $w(\rho) $ is the width of $\rho$ defined in~\eqref{eq:width}.
Therefore,
 \begin{equation*}
   \Delta(Y_{0:T})   \leq \frac{\epsilon^2}{2I\sigma^2} w(\rho)\E_{\mathcal Q_0}[\tilde M_{\mathcal I}].\label{ineq:bandit1}
 \end{equation*}
 By Pinsker’s inequality~\cite[Lemma 11.6.1]{cover1999elements}, we have 
			\begin{align*}
				d^{\tilde{\mathcal F}}_{\mathrm{TV}} (\mathcal Q_0, \mathcal Q_{\mathcal I})&=\sup_{A\in \tilde{\mathcal F}} | \mathcal Q_0 (A) - \mathcal Q_{\mathcal I}(A)  |  \nonumber\\*
    &\leq \frac{\epsilon}{2\sigma\sqrt{I}} \sqrt{w(\rho) \E_{\mathcal Q_0}[\tilde M_{\mathcal I}]}.
			\end{align*}
\end{IEEEproof}

\yan{It may be possible to improve Lemma~\ref{lm:partswitch1} to  obtain a tighter lower bound.
	In the proof of Lemma~\ref{lm:partswitch1}, there is an inequality $|N_t| \leq N_t'$ used in~\eqref{ineq:ntandntprime}, where
 $N_t=\langle \tilde A_{\rho(t)}, \mathcal I\rangle - \langle \tilde A_{t}, \mathcal I \rangle $ is  the difference between numbers of arms in $\mc I$ played at time $\rho(t)$ and $t$ and	$ N_t'\triangleq\langle \tilde A_{\rho(t)}\oplus \tilde  A_{t}, \mathcal I\rangle $ is the total number of switches the player performs to or from base arms in $ \mc I $ between  rounds $ \rho(t) $ and $ t $. It is easy to observe that in many cases of $\tilde  A_{\rho(t)}$ and $\tilde A_t$, this inequality is not tight and could even be quite  loose. Therefore, the design of a loss sequence that tightens this inequality  is a good future research direction to possibly obtain a tighter lower bound.}

\subsection{Proof of Theorem~\ref{thm:lowerbound-semi}: Semi-bandit Feedback}\label{subsec:lowerbound-semi}
\begin{algorithm}[t]
	\SetKwInOut{Input}{Input}
	\SetKwInOut{Output}{Output}
	\SetAlgoLined
	\Input{Time horizon $T$, number of actions $K$ and combinatorial arm size $I$}  
	\begin{enumerate}
		\item[\emph{Step 1:}] Set $ \sigma =\frac{1}{(9 \log_2 T)} $ and $ \epsilon = \frac{(\lambda K)^{\frac{1}{3}}I^{-\frac{2}{3}}T^{-\frac{1}{3}}}{9\log_2 T} $.
 Choose $ \chi \in \mc A$ uniformly at random and then generate $W_t^i$ \\
 for $t\in [T]$ and $i\in [K]$ according to~\eqref{eq:wdefsemi}.
		\item[\emph{Step 2:}] For  $ \forall\,t\in [T] $ and $ \forall \,x\in [K] $, set $x$-th components
  of $\tilde L_t$ and $L_t$ as
 \begin{minipage}{0.42\textwidth}
\begin{equation}\label{eq:Lprime2}
    \tilde{L}_{tx} = W_t^x +\frac{1}{2} - \epsilon \chi_x,\quad L_{tx} = \text{clip} (\tilde{L}_{tx}).
\end{equation}
\end{minipage}
	\end{enumerate}
	\Output{Loss sequence $L_{1:T}$.}	
	\caption{The Combinatorial Diverse-Noise (CDN) loss sequence}\label{alg:sequence2}
\end{algorithm}



Similar to the analysis of lower bound under bandit feedback in~\S\ref{subsec:lowerboundbandit}, we will prove  Theorem~\ref{thm:lowerbound-semi} by constructing a stochastic loss sequence  $L_{1:T}$ in Algorithm~\ref{alg:sequence2}.
Let $\rho(t)$ be defined according to~\eqref{eq:rhodef}.
For any $i\in [K]$, define $ W_{t}^i$ for $ t\in [T] $ and  recursively by $W_0^i=0$ and
\begin{align}\label{eq:wdefsemi}
	W^i_t &= W^i_{\rho(t)}+\xi^i_t, \quad 	\forall\,  t\in [T],
\end{align}
where $\xi^i_t$, $t\in [T], i\in [K]$ are independent zero-mean, variance $ \sigma^2 $ Gaussian variables.  The difference between the loss sequences in Algorithm~\ref{alg:sequence2} and Algorithm~\ref{alg:sequence1} lies in the independent and identically distributed (i.i.d.) nature of the added Gaussian noises $\xi_t^i$ for each arm $i\in [K]$.
This modification effectively tackles the challenge presented by the semi-bandit feedback scenario, where all losses for each base arm in the selected combinatorial arm are observed. 
Under the loss sequence $\tilde{L}_{1:T}$ in Algorithm~\ref{alg:sequence1} and semi-bandit feedback, the observed losses for each base arm in the chosen combinatorial arm at time $t\in [T]$ when $\chi =\emptyset$ are all the same while the losses for each base arm when $\chi \in \mc A$ may not be the same and differ by $\epsilon$ (see definition of $\epsilon$ in Algorithm~\ref{alg:sequence1}). 
Then the supports for observed losses under the adversaries in Algorithm~\ref{alg:sequence1} when $\chi = \emptyset$ and $\chi \in \mc A$ are different and thus the KL divergence on the observed losses under two adversaries is infinite. To overcome this issue, the loss sequence in Algorithm~\ref{alg:sequence2} is introduced, which induces the same support for the observed losses under both $\chi =\emptyset$ and $\chi \in \mc A$. Further details will be provided in the proof of the following Lemma~\ref{lm:lowerbound2}.

  By Yao's minimax principle, Theorem~\ref{thm:lowerbound-semi} can be proved if the following lemma is verified.
\begin{lemma}\label{lm:lowerbound2}
	Consider the combinatorial bandit problem with switching costs under the semi-bandit feedback.
	Let $L_{1:T}$ be the stochastic sequence of loss functions defined in Algorithm~\ref{alg:sequence2}.
	When $K\geq 3I$ and $ T\geq \max \{\frac{\lambda K}{I^2},6\} $, for any deterministic player $\pi$, we have 
	$$	R_{\lambda}(\pi, L_{1:T})\geq \frac{(\lambda K I)^{\frac{1}{3}} T^{\frac{2}{3}}}{60\log_2 T}. $$
\end{lemma}
   To prove Lemma~\ref{lm:lowerbound2},    we need to analyze the expected regret under an arbitrary deterministic policy $\pi: [0,1]^{I\times T}\to \mc A^T$  when the loss sequence is $L_{1:T}$. Note that under $ L_{1:T}$,  a deterministic policy $\pi$ yields an action sequence $ A_{1:T} \in \mc A^T$ so that $A_t$ is a function of the player's past observations $Z_{1:t-1} $ with $Z_t  =  A_t\circ  L_t $.  
Similar to~\S\ref{subsec:lowerboundbandit},
  we first analyze the regret under the loss sequence $\tilde{L}_{1:T}$ defined in Algorithm~\ref{alg:sequence2} that the player would suffer on the deterministic policy $\pi\, \circ \,$clip. The policy $\pi\, \circ \,$clip yields the same
  action sequence $A_{1:T}$ with that of $L_{1:T}$ under $\tilde L_{1:T}$. Thus we only need to analyze the regret under the loss sequence $\tilde{L}_{1:T}$ and the action sequence $A_{1:T}$.
		Let 
  \begin{equation}\label{eq:defyseq}
      Y_0=\frac{1}{2},\ Y_t = A_t \circ \tilde L_t.
  \end{equation}
  and $Y_{t,j}=\tilde{L}_{t,i} A_{t,i}$. Define $ Y_S= \{Y_t\}_{t\in S} $ and let $ \Delta (Y_S| Y_{S'}) $ be the KL divergence between the distribution of $ Y_S $ conditioned on $ Y_{S'} $ under $ \mathcal Q_0$ and $ \mathcal Q_{\mathcal I} $ as defined in~\eqref{eq:defKL}, where $ \mathcal Q_0$ and $ \mathcal Q_{\mathcal I} $ are defined as in~\eqref{eq:Qdef1} and~\eqref{eq:Qdef2} under the loss sequence $\tilde{L}_{1:T}$ in Algorithm~\ref{alg:sequence2}.

  In the following lemma, we derive a relation between $\Delta(Y_{0:T})$ and $M_{\mc I}$, where $M_{\mc I}$ is the total number of switches the player performs to or from base arms in $ \mathcal I $ during the whole time horizon and defined as~\eqref{eq:defMI} in~\S\ref{subsec:lowerboundbandit}.
  Based on the inequality in the following Lemma~\ref{lm:lowerboundsemi}, we can prove Lemma~\ref{lm:lowerbound2} by the similar verification with~\cite{dekel2014bandits} and the whole proof is detailed in Appendix~\ref{app:semibanditlower}.
  \begin{lemma}\label{lm:lowerboundsemi}
  Under the loss sequence $\tilde{L}_{1:T} $, which is defined in~\eqref{eq:Lprime2} of Algorithm~\ref{alg:sequence2} with $\chi = \mc I\in \mc A$,  it holds that $\Delta(Y_{0:T})  \leq \frac{\epsilon^2}{2\sigma^2} w(\rho) \E_{\mathcal Q_0}[M_{\mathcal I}]$.
  \end{lemma}
  \begin{IEEEproof}
Using the chain rule,
\begin{align*}
	\Delta(Y_{0:T}) \triangleq	\Delta(Y_{0:T} | \emptyset) 
	&=\sum_{t=1}^T   \Delta (Y_t |Y_{\mc S(t)}),
\end{align*}
where $\mc S(t)$ is defined as in~\eqref{eq:rhostar}.
Since $Y_{t,j}$ are independent for different $j\in [K]$ given $Y_{\mc S(t)}$, 
\begin{align*}
	\Delta (Y_t |Y_{\mc S(t)})&=\sum_{i: A_{t,i}=1} \Delta (Y_{t,i}|Y_{\mc S(t)}).
\end{align*}
Let $\mc I_i$ denote the $i$-th component of $\mc I$. In the following, we analyze the value of $\Delta (Y_{t,i}|Y_{\mc S(t)})$ for $t\in [T]$ and $i\in[K]$ such that $A_{t,i}=1.$
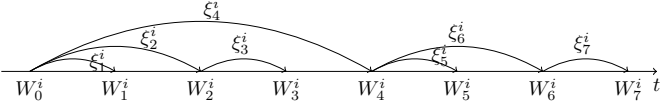
\begin{figure}[!t]
\centering
\resizebox{\columnwidth}{!}{%
\begin{tikzpicture}
\path (0.5, 0) coordinate (A) node[below] {$W_0^i$};
\path (2, 0) coordinate (B) node[below] {$W_1^i$};
\path (3.5, 0) coordinate (C) node[below] {$W_2^i$};
\path (5, 0) coordinate (D) node[below] {$W_3^i$};
\path (6.5, 0) coordinate (E) node[below] {$W_4^i$};
\path (8, 0) coordinate (F) node[below] {$W_5^i$};
\path (9.5, 0) coordinate (G) node[below] {$W_6^i$};
\path (11, 0) coordinate (H) node[below] {$W_7^i$};
\path (11.5, 0) coordinate (I) node[below] {$t$};
\path (3.7, 1.4) node[below] {$\xi_4^i$};
\path (8, 1) node[below] {$\xi_6^i$};
\path (10.2, 0.8) node[below] {$\xi_7^i$};
\path (1.7, 0.5) node[below] {$\xi_1^i$};
\path (2.6, 0.9) node[below] {$\xi_2^i$};
\path (4.2, 0.8) node[below] {$\xi_3^i$};
\path (7.7, 0.6) node[below] {$\xi_5^i$};
\draw (0,0) edge[->] (11.5,0);
    \path[->] (A) edge [bend left] node [right] {} (B);
     \path[->] (A) edge [bend left] node [right] {} (C);
     \path[->] (C) edge [bend left] node [right] {} (D);
     \path[->] (A) edge [bend left] node [right] {} (E);
     \path[->] (E) edge [bend left] node [right] {} (F);
     \path[->] (E) edge [bend left] node [right] {} (G);
     \path[->] (G) edge [bend left] node [right] {} (H);
\end{tikzpicture}}
\caption{An illustration of the definition of $W_t^i$ for $i\in [K]$. The value of $W_t^i$ is obtained by summing the i.i.d. Gaussian variables $\xi_{t'}^i$'s on the edges along the path from $W_0^i,$ i.e. 
 summing over all $t'\in \mc S(t) \cup \{t\} \setminus \{0\}$. For example, $W_7^i=$ $ \xi_4^i +\xi_6^i+\xi_7^i$.}\label{fig:semi_illustration}
\end{figure}
\begin{enumerate}
	\item  When  $\mc I_i=0$, 	the distributions of $Y_{t,i}$ conditioned on $Y_{\mc S(t)}$ are the same under both $\mc Q_0$ and $\mc Q_{\mc I}.$ Specifically, when $i$ is not the optimal arm, the probability distributions of $Y_{t,i}= \tilde{L}_{t,i}$ conditioned on $Y_{\mc S(t)}$ under  $\mc Q_0$ and $\mc Q_\mc I$ are the same since the definitions of $\tilde{L}_{t,i}$ in Algorithm~\ref{alg:sequence2} are the same under $\mc Q_0$ and $\mc Q_\mc I$ when $i\notin \mc I$. 
 Thus $\Delta (Y_{t,i}|Y_{\mc S(t)})=0.$ 
\item When $\mc I_i=1$ and $A_{t',i}=0$  for all $ t'\in \mc S(t)$, 	the distributions of $Y_{t,i}$ conditioned on $Y_{\mc S(t)}$ are  $N\big(\frac{1}{2}, |\mc S(t)| \sigma^2\big)$ under  $\mc Q_0$ and  $N\big(\frac{1}{2}-\epsilon, |\mc S(t)| \sigma^2\big)$ under $\mc Q_{\mc I}$. Then $$\Delta (Y_{t,i}|Y_{\mc S(t)}) = d_{\mathrm{KL}}\big(N(0,|\mc S(t)|\sigma^2)\big\rVert N (\epsilon, |\mc S(t)| \sigma^2)\big).$$ For example, suppose $t=7$ and $A_{4,i}=A_{6,i}=0,$ then we have $W_7^i =\frac{1}{2}+\xi_4^i+\xi_6^i+\xi_7^i$ under $\mc Q_0$ and $W_7^i =\frac{1}{2}-\epsilon+\xi_4^i+\xi_6^i+\xi_7^i$ under $\mc Q_\mc I$ (see the illustration in Figure~\ref{fig:semi_illustration}). Thus $W_7^i\sim N(\frac{1}{2}, 3\sigma^2)$ under $\mc Q_0$ and $W_7^i\sim N(\frac{1}{2}-\epsilon, 3\sigma^2)$ under $\mc Q_\mc I$.
	\item When $A_{t',i}=1$ for some $t'\in \mc S(t)$ and $A_{r,i}=0$ for $r\in \mc S(t)$ with $r>t'$,
	the distributions of $Y_{t,i}$ conditioned on $Y_{\mc S(t)}$ are both  $N(Y_{t' ,i}, c\sigma^2)$ under $\mc Q_0$ and $\mc Q_{\mc I}$, where $c=\big|\mc S(t)\cup \{t\}\setminus \{0,\dots, t'\}\big|$. Then $\Delta (Y_{t,i}|Y_{\mc S(t)})=0.$
	For example, suppose $t=7$, $A_{4i} =1$ and $A_{6i}=0,$ then we have $W_7^i =Y_{4i}+\xi_6^i+\xi_7^i$ (see the illustration in Figure~\ref{fig:semi_illustration}) and thus $W_7^i\sim N(Y_{4i}, 2\sigma^2)$ when $Y_{4i}$ has been observed.

\end{enumerate}
			Therefore, by setting $$N^*_t =  \big| \{j\in [K]:A_{t,j} =\mc I_j=1, A_{t',j}=0,\, \forall\, t'\in \mc S(t)\}\big|,$$ we have
\begin{align*}
	&\Delta (Y_t |Y_{\mc S(t)}) \nonumber\\*
 &= \sum_{i=1}^I \mathcal Q_0 (N^*_t =i) id_{\mathrm{KL}}\big (N (0, |\mc S(t)| \sigma^2 )\big\rVert N (\epsilon, |\mc S(t)| \sigma^2 ) \big)\\
	&= \sum_{i=1}^I \mathcal Q_0 (N^*_t =i)   \frac{i\epsilon^2}{2|\mc S(t)|\sigma^2}.
\end{align*}
We observe that $ \{N^*_t =i \}$ is the event that the player switched at least $ i $ arms to actions in $ \mathcal I $ between rounds $ \rho(t) $ and $ t $. Then
\begin{align}
	\Delta(Y_{0:T}) &= \frac{\epsilon^2}{2\sigma^2} \sum_{t=1}^T   \frac{1}{|\mc S(t)|} \sum_{i=1}^I 	Q_0 (N^*_t=i)  i \nonumber\\
	&\leq  \frac{\epsilon^2}{2\sigma^2} \sum_{t=1}^T   \E_{Q_0} [N^*_t]\nonumber\\
	&\leq  \frac{\epsilon^2}{2\sigma^2} \sum_{t=1}^T   \E_{Q_0} [N'_t]\label{eq:NprimeNstr}\\
	&  \leq \frac{\epsilon^2}{2\sigma^2} w(\rho) \E_{\mathcal Q_0}[M_{\mathcal I}],\nonumber
\end{align}
where $ N_t'\triangleq\langle \tilde A_{\rho(t)}\oplus \tilde A_{t}, \mathcal I\rangle $ is the total number of switches the player performs to or from base arms in $ \mc I $ between rounds $ \rho(t) $ and $ t $, and the last inequality holds due to~\eqref{eq:banditswitch}.\end{IEEEproof}
 \yan{  It may also be possible to improve Lemma 8 to obtain a tighter lower bound. 	In the proof of Lemma 8, there is an inequality $N_t^* \leq N_t'$ used in~\eqref{eq:NprimeNstr}, where
 $$N^*_t =  \big| \{j\in [K]:A_{t,j} =\mc I_j=1, A_{t',j}=0,~\forall\,  t'\in \mc S(t) \}\big|,$$ and $ N_t'\triangleq\langle A_{\rho(t)}\oplus A_{t}, \mathcal I\rangle $.  It is clear that for many cases of $A_{\rho(t)}$ and $A_t$, this inequality is not tight and could even be quite loose. Therefore, the design of a loss sequence that tightens this inequality constitutes a good future research direction to possibly obtain a tighter lower bound.}
\section{Algorithm for Bandit Feedback and Semi-bandit Feedback}\label{sec:banditsemi}
\begin{figure}[!t]
\centering
\resizebox{\columnwidth}{!}{%
\begin{tikzpicture}
\path (0.5, 0) coordinate (A) node[below] {$t=0$};
\path (10, 0) coordinate (B) node[below] {$t$};
\draw[arrows={->[slant=.3]}]  (0,0) -- (10,0);
\draw[thick, blue, arrows={->[slant=.3]}] (1.75,-0.75)--(2.5,0);
\draw[thick, blue, arrows={->[slant=.3]}] (3.75,-0.75)--(4.5,0);
\draw[thick, blue, arrows={->[slant=.3]}] (5.75,-0.75)--(6.5,0);
\draw[thick, blue, arrows={->[slant=.3]}] (7.75,-0.75)--(8.5,0);
\draw[thick] (0.5,-0.1)--(0.5,0.1);
\draw[thick] (2.5,-0.1)--(2.5,0.1);
\draw[thick] (4.5,-0.1)--(4.5,0.1);
\draw[thick] (6.5,-0.1)--(6.5,0.1);
\draw[thick] (8.5,-0.1)--(8.5,0.1);
\path (1.5, 0) coordinate (A1) node[above] {$B_1$};
\path (3.5, 0) coordinate (A2) node[above] {$B_2$};
\path (5.5, 0) coordinate (A1) node[above] {$B_3$};
\path (7.5, 0) coordinate (A2) node[above] {$B_4$};
\path (1.75,-0.75) node[below, red] {Switch};
\path (3.75,-0.75) node[below, red] {Switch};
\path (5.75,-0.75) node[below, red] {Switch};
\path (7.75,-0.75) node[below, red] {Switch};
  
\end{tikzpicture}}
\caption{An illustration of the batches in algorithm. The whole time horizon are divided into batches and during each batch, the player does not change the choice of the combinatorial arm.}\label{fig:block_alg}
\end{figure}
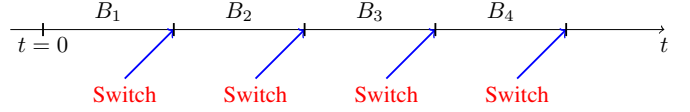

In this section, we will introduce our algorithms for the two types of feedback.
We will use the batched algorithm to restrict the number of switches between actions by dividing the whole time horizon into batches and forcing the algorithm to play the same action for all the rounds within a batch as shown in Figure~\ref{fig:block_alg}.

\subsection{Algorithm for Bandit Feedback}
The Exp2 with John's exploration algorithm~\cite{bubeck2011introduction} is an efficient algorithm for the combinatorial bandit problem under bandit feedback.  In Algorithm~\ref{alg:EXP2John}, we introduce a refinement of this algorithm, called \textsc{Batched-Exp2} with John's exploration to take into account switching costs, where John's exploration distribution can be obtained in~\cite[\S~7.3.2]{bubeck2011introduction}. In this section, we prove the following theorem, which is a formal version of Theorem~\ref{thm:exp2informal}, to obtain a bound for the regret of the proposed algorithm.

\begin{theorem}\label{thm:exp2}
    	Let  $\pi$ be the policy of \textsc{Batched-Exp2} with John's exploration distribution. The time horizon $T$ is divided into $N$ batches with lengths satisfying $B_n = B = \left \lceil{\lambda^{\frac{2}{3}}K^{-\frac{1}{3}}T^{\frac{1}{3}}I^{-\frac{1}{3}}}\right \rceil 
$ for $1\leq n\leq \left \lfloor  \frac{T}{B}\right \rfloor$ and $B_N = T-(N-1)B$ with $N = \left \lfloor  \frac{T}{B}\right \rfloor+1$. Let $\gamma = \eta BI K$ and $\eta =\sqrt{\frac{\ln \binom{K}{I}}{3NK(BI)^2}}$. Then for any adversary $l_{1:T}\in \mc L$, the $\lambda$-switching regret satisfies
\begin{equation*}
    R_{\lambda}(\pi,l_{1:T}) =O  \big((\lambda K)^{\frac{1}{3}} T^{\frac{2}{3}}I^{\frac{4}{3}} \big).
\end{equation*}
\end{theorem}

   \begin{algorithm}[t]
	\SetKwInOut{Input}{Input}\SetKwInOut{Output}{output}
	\SetAlgoLined
	\Input{John's exploration distribution $ \mu $ over~$\mc A$; \\
 batch lengths $ B_1,\dots, B_N $ s.t.\  $\sum_{i=1}^N B_n=T$; \\
 mixing coeff.\ $ \gamma\! \in\! (0,1)$ and 
learning rate~$\eta$;\\
   $q_1 =$$ (\frac{1}{|\mc A|}, \dots, \frac{1}{|\mc A|}) \in \mathbb{R}^{|\mc A|}$.} 
		\For{$1\leq n\leq N$}{
   \begin{minipage}{0.44\textwidth}
			\begin{enumerate}
				\item[(a)] Let $ p_{n}=(1-\gamma)q_{n}+\gamma \mu $, and select a combinatorial arm $A(n)$ with respect to $p_n$. Pull the selected combinatorial arm for $B_n$ times, which
				 then incurs a loss $ X(n) =\big\langle A(n),\ l(n)\big\rangle $, where $l(n)=\sum_{b\in [B_n]} l(n,b)$ and $ l(n,b) $ is  
     the loss vector at the $ b $-th time due to the pulling of $A(n)$  in this batch. 
				\item[(b)]    Estimate the loss vector $l(n)$ by $ \tilde{l}(n)=X(n)\Sigma^+_{n-1}A(n) $, with $ \Sigma_{n-1} = \E_{A\sim p_n}[AA^T  ] $ where $ \Sigma^+_{n-1} $ is the pseudo-inverse of $ \Sigma_{n-1} $.
    \item[(c)] 
    Update the exponential weights. That is, for all $A\in \mc A$,
    \begin{equation*}
    q_{n+1}(A) = \frac{q_{n}(A)\exp\big(-\eta \langle A, \tilde{l}(n)\rangle \big) }{\sum_{A'\in \mc A} q_{n}(A')\exp\big(-\eta \langle A', \tilde{l}(n)\rangle \big)}.
    \end{equation*}
	\end{enumerate}
		\end{minipage}
  }
	\caption{\textsc{Batched-Exp2} with John's exploration}\label{alg:EXP2John}
\end{algorithm}
\begin{IEEEproof}[Proof of Theorem~\ref{thm:exp2}]
By definition, we have $l(n) = \sum_{b=1}^{B_n}l_{(n-1)B+b} \in [0,B]$.
Since $A(n)\in \mc A$, the accumulated loss in each batch satisfies
\begin{equation*}
    X(n) =\big\langle A(n), l(n)\big\rangle \leq BI.
\end{equation*}
    By~\cite[Theorem $7.6$]{bubeck2011introduction}, when $\gamma = \eta BI K$ and $\eta =\sqrt{\frac{\ln |\mc A|}{3NK(BI)^2}}$, the pseudo-regret satisfies
    \begin{align*}
        R(\pi, l_{1:T})& \leq 2BI \sqrt{3NK\ln |\mc A|} \\
        &\leq 4\lambda^{\frac{2}{3}}K^{-\frac{1}{3}}T^{\frac{1}{3}}I^{-\frac{1}{3}}I\sqrt{6 \lambda^{-\frac{2}{3}}K^{\frac{1}{3}}T^{\frac{2}{3}}I^{\frac{1}{3}}K I\ln \frac{eK}{I}}\\
        &= 4\sqrt{6\ln \frac{eK}{I}} \lambda^{\frac{1}{3}} K^{\frac{1}{3}} T^{\frac{2}{3}}I^{\frac{4}{3}}.
    \end{align*}
    Thus 
    \begin{align*}
          R_{\lambda}(\pi, l_{1:T}) & \leq R_T(\pi, l_{1:T}) + \lambda I N  \\
          & \leq 4\sqrt{6\ln \frac{eK}{I}} \lambda^{\frac{1}{3}} K^{\frac{1}{3}} T^{\frac{2}{3}}I^{\frac{4}{3}} +2\lambda^{\frac{1}{3}} K^{\frac{1}{3}} T^{\frac{2}{3}}I^{\frac{4}{3}} \\
          &= \Big(4\sqrt{6\ln \frac{eK}{I}}+2\Big)\lambda^{\frac{1}{3}} K^{\frac{1}{3}} T^{\frac{2}{3}}I^{\frac{4}{3}}.
    \end{align*}
\end{IEEEproof}
 
\subsection{Algorithm for Semi-Bandit Feedback}

		  \begin{algorithm}[t]
	\SetKwInOut{Input}{Input}\SetKwInOut{Output}{output}\SetKwInOut{Define}{Define}
	\SetAlgoLined
 \Define{$F_n(a)=\frac{1}{\eta_n} \sum_{i=1}^K \ln \frac{1}{a_i} .$}
	\Input{$\eta_1=\eta$; \\
 batch lengths $ B_1,\dots, B_N $ s.t. $\sum_{i=1}^N B_n=T$;\\  $n=1, N_0=0.$ }
	
 \For{$\beta=0,1,\dots $}{
		$ a'_n=\argmin_{a\in\mathrm{Co}(\mc A)} F_1(a)$.\\
  		\While{$n\leq N$}{
     \begin{minipage}{0.405\textwidth}
\begin{enumerate}		
  \item
  $a_n = \argmin_{a\in  \mathrm{Co}(\mc A)} \{D_{F_n}(a,a_n')  
 \}$, where $D_{F_n}$ is defined as in~\eqref{eq:semiDF}.
\item Sample $A(n)$ such that $ \E[A(n)]=a_n$ and then pull it for $ B_n $ times.
				Observe $A(n) \circ l(n,b)$   for $b\in [B_n]$
				and
				incur a loss $$ X(n) =  \langle A(n) , \; l(n) \rangle ,$$  where $l(n)=\sum_{b\in [B_n]} l(n,b)$ with $ l(n,b) $  being  the loss vector   at $ b $-th 
    time of pulling $A(n)$  in this	batch.
\item    Compute the estimator $\hat l(n)$ with $$\big(\hat{l}(n)\big)_i =\frac{ \big(A(n)\big)_i\big(l(n)\big)_i}{(a_{n})_i},$$ for $i\in [K]$, where we use $(v)_i$ to 
denote the $i$th component in $v.$
\item    Update $$a_{n+1}' =\argmin_{a\in \mathrm{co}(\mc A)}  \langle a ,  \hat{l}(n) \rangle + D_{F_{n}}(a, a_n').$$
\end{enumerate}
\end{minipage}
\If{
$\sum_{s=N_{\beta}+1}^n \lVert A(n) \circ l(n) \rVert_2^2\geq \frac{K\ln T}{3\eta^2}$
}{
$\eta_{n+1} \leftarrow \eta_n/2$, $N_{\beta+1}\leftarrow n$, $n\leftarrow n+1$\;
\textbf{break}\;
}
$\eta_{n+1}\leftarrow \eta_n$. $n\leftarrow n+1.$ 
}
}
\caption{\textsc{Batched-BROAD}}\label{alg:BROAD}
\end{algorithm}
We propose the \textsc{Batched-BROAD} algorithm as stated in Algorithm~\ref{alg:BROAD}, based on the BROAD algorithm in~\cite{wei2018more}, which is an Online Mirror Descent algorithm with log-barrier regularizer. For a regularizer $F:\mathbb R^K\to \mathbb R$, define 
\begin{align}
D_F(p,q) \triangleq F(p)-F(q) -  \langle \nabla F(q) , \; p-q \rangle. \label{eq:semiDF}
\end{align}
In the following theorem, we first prove \textsc{Batched-BROAD} can achieve a $\lambda$-switching regret as shown in Theorem~\ref{thm:broadinformal}.
\begin{theorem}\label{thm:broad}
    Let  $\pi $ be the policy of \textsc{Batched-BROAD}. The time horizon $T$ is divided into $N$ batches with lengths satisfying   $B_n =B=\left \lfloor (TI)^{\frac{1}{3}}\lambda^{\frac{2}{3}}K^{-\frac{1}{3}}+1
	\right \rfloor$, for $n = 1,\dots, N-1$ and $B_N=T-\sum_{n=1}^{N-1}B_n$ with $N=\left \lfloor \frac{T}{B} \right \rfloor+1$. Let $\eta=\min \{ \frac{1}{18IB^2},\frac{1}{81}\}$. Then for any adversary $l_{1:T}\in \mc L$, when $T\geq \frac{K}{I\lambda^2}$  the $\lambda$-switching regret satisfies
    \begin{equation*}
        R_{\lambda}(\pi,l_{1:T})=\tilde O\big( (\lambda K)^{\frac{1}{3}}(TI)^{\frac{2}{3}}+KI\big).
    \end{equation*}
\end{theorem}
 In the following lemma, we first give the generalized analysis of BROAD algorithm for losses in varying intervals over time. Based on the lemma, we can then prove Theorem~\ref{thm:broad}.
  
\begin{lemma}\label{lm:broad}
Consider a general combinatorial multi-armed bandit problem with semi-bandit feedback where $l_t\in [0,b_t]^K$ and $\mc A=\{A\in \{0,1\}^K: \lVert A \rVert_1 =I \}$. 
   Let the agent's policy $\pi$ be obtained by Algorithm~\ref{alg:BROAD} with batch lengths $B_n=1$ for all $n\leq N$.   If for all $t\leq T$, $\eta_t  \leq \min\{\frac{1}{18Ib_t^2}, \frac{1}{81}\}$,  then the pseudo-regret satisfies
    \begin{equation*}
        R(\pi, l_{1:T}) = O\Bigg(\sqrt{(KI\ln T )\sum_{t=1}^Tb_t^2} +KI\ln T\Bigg)
    \end{equation*}
\end{lemma}
\begin{IEEEproof}
    For the combinatorial arm $A_t$ pulled at time $t$  and the loss vector $l_t\in [0,b_t]^{K}$ at time $t$, we have
    \begin{equation*}
         \lVert A_t \circ l_t \rVert_2^2= \sum_{i:A_{t,i}=1}l_{t,i}^2\leq Ib_t^2.
    \end{equation*}
    Then we have $\eta_t  \lVert A_t \circ l_t \rVert_2^2\leq \frac{1}{18} $ and thus by~\cite[Theorem $8$]{wei2018more}, we have 
     \begin{equation*}
        R(\pi, l_{1:T}) = O\Bigg(\E \Bigg[\sqrt{(K\ln T) \sum_{t=1}^T\lVert A_t \circ l_t \rVert_2^2 +KI\ln T}\Bigg]\Bigg),
    \end{equation*}
    where the expectation is taken over the randomized choice of $A_t.$
Since $ \lVert A_t \circ l_t \rVert_2^2\leq Ib_t^2$, the proof is completed.
\end{IEEEproof}
\begin{IEEEproof}[Proof of Theorem~\ref{thm:broad}]
In $n$-th batch, the accumulated loss vector $l(n)\in [0,B_n]^K$.
By Lemma~\ref{lm:broad}, when $T\geq \frac{K}{I\lambda^2}$ the pseudo-regret satisfies
 \begin{align*}
        R(\pi, l_{1:T}) &= O\Bigg(\sqrt{(KI\ln T )\sum_{n=1}^NB_n^2} +KI\ln T\Bigg)\\
        &\leq O\bigg(\sqrt{(2KI\ln T )T(TI)^{\frac{1}{3}}\lambda^{\frac{2}{3}}K^{-\frac{1}{3}}} +KI\ln T\bigg)  \\
        &=O\bigg(\sqrt{2\ln T } (\lambda K)^{\frac{1}{3}}(TI)^{\frac{2}{3}}+KI\ln T\bigg) .
    \end{align*}
 Since $N\leq \frac{T}{(TI)^{\frac{1}{3}}\lambda^{\frac{2}{3}}K^{-\frac{1}{3}}} +1 $, the $\lambda$-switching regret satisfies
    \begin{align*}
        R_\lambda (\pi, l_{1:T})& \leq R(\pi,l_{1:T})+ \lambda I N \nonumber\\*
        &=O(\sqrt{\ln T } (\lambda K)^{\frac{1}{3}}(TI)^{\frac{2}{3}}+KI\ln T).
    \end{align*}
\end{IEEEproof}
\section{Numerical Results}\label{sec:numerical}
  \begin{figure*}[t]
	\centering
	\begin{subfigure}{0.32\textwidth}
		\includegraphics[width=\textwidth]{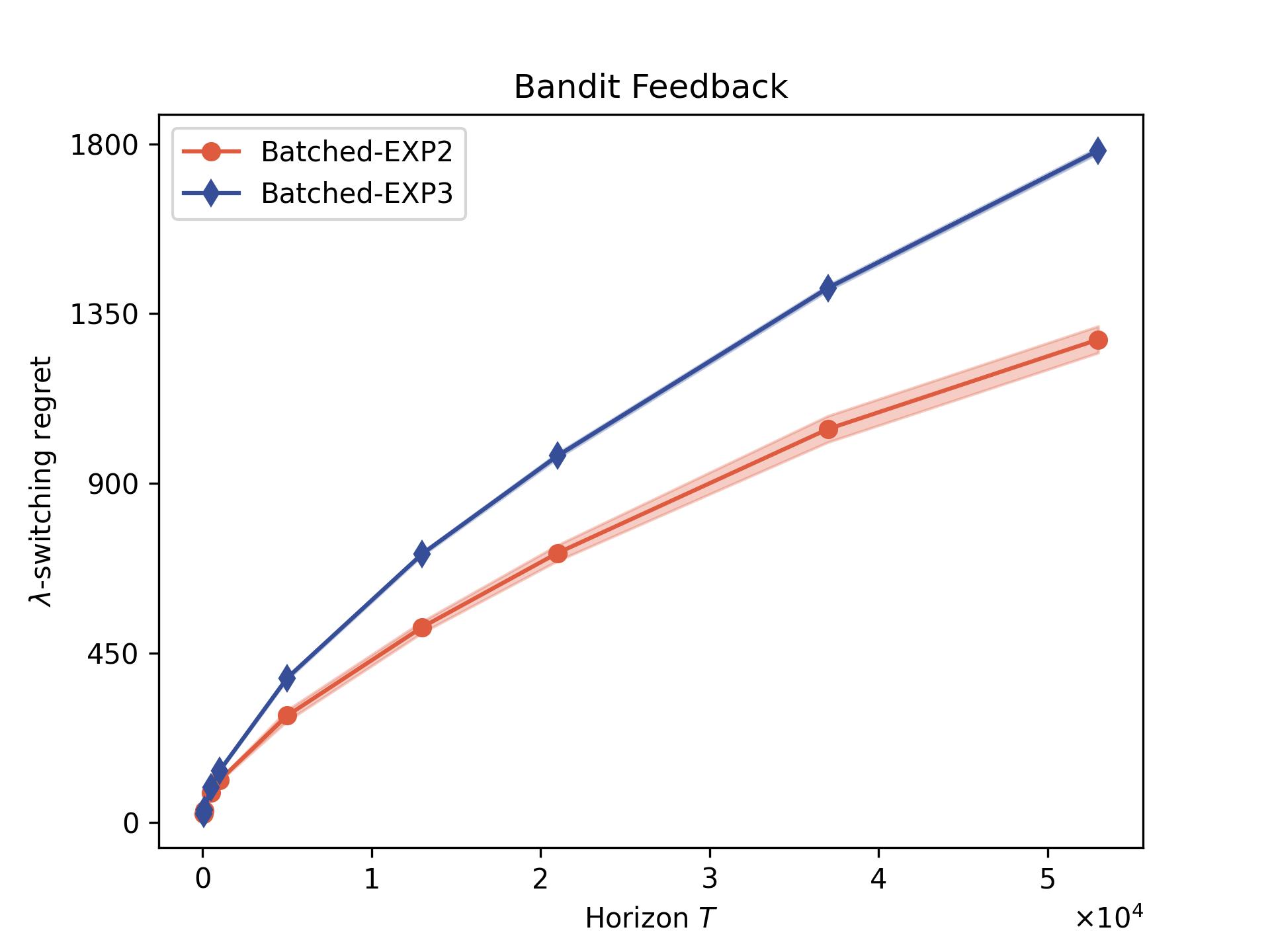}
		\caption{ $K=10$, $I=3$ and $\lambda = 1$ under  bandit feedback using the CIN loss sequence. }
		\label{fig:figure1}
	\end{subfigure}
 \hfill
  	\begin{subfigure}{0.32\textwidth}
		\includegraphics[width=\textwidth]{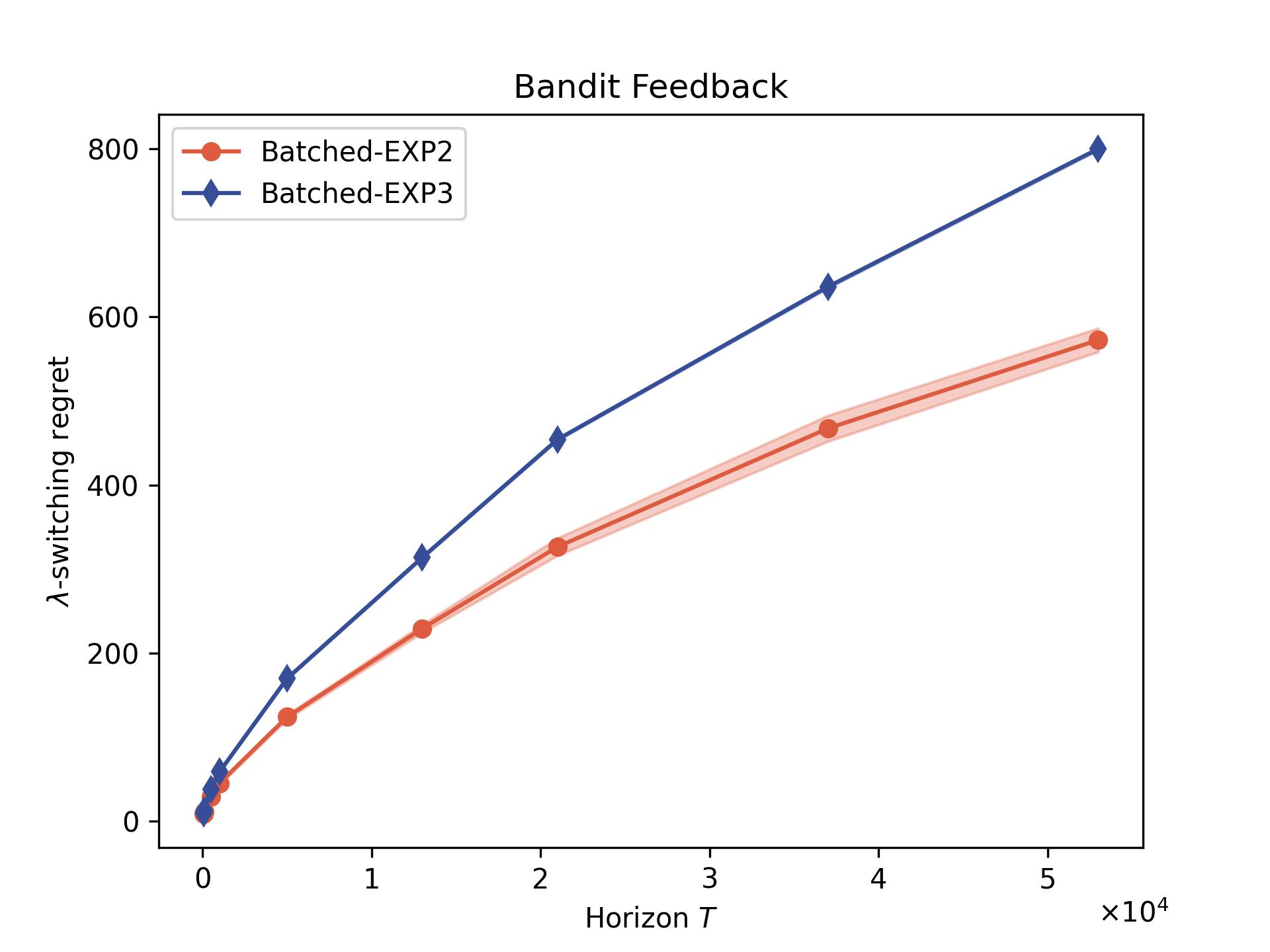}
		\caption{ $K=10$, $I=3$ and $\lambda = 0.1$ under  bandit feedback using the CIN loss sequence. }
		\label{fig:figure2a}
	\end{subfigure}
\hfill
\begin{subfigure}{0.32\textwidth}
		\includegraphics[width=\textwidth]{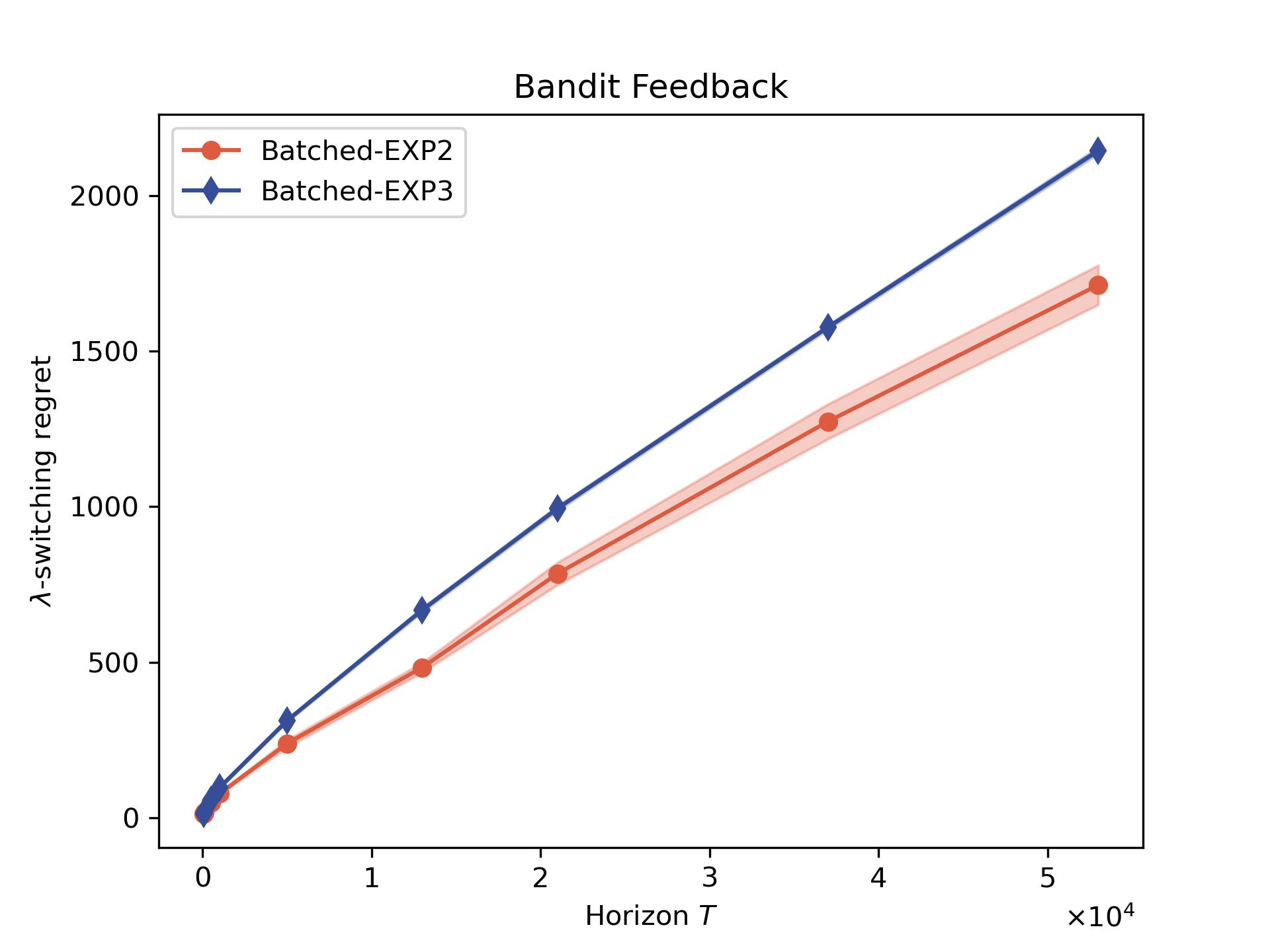}
		\caption{ $K=10$, $I=3$ and $\lambda = 1$ under  bandit feedback using the SC$(0.01)$ adversary. }
		\label{fig:banditcomparels1.0}
	\end{subfigure}
	\hfill
  	\begin{subfigure}{0.32\textwidth}
		\includegraphics[width=\textwidth]{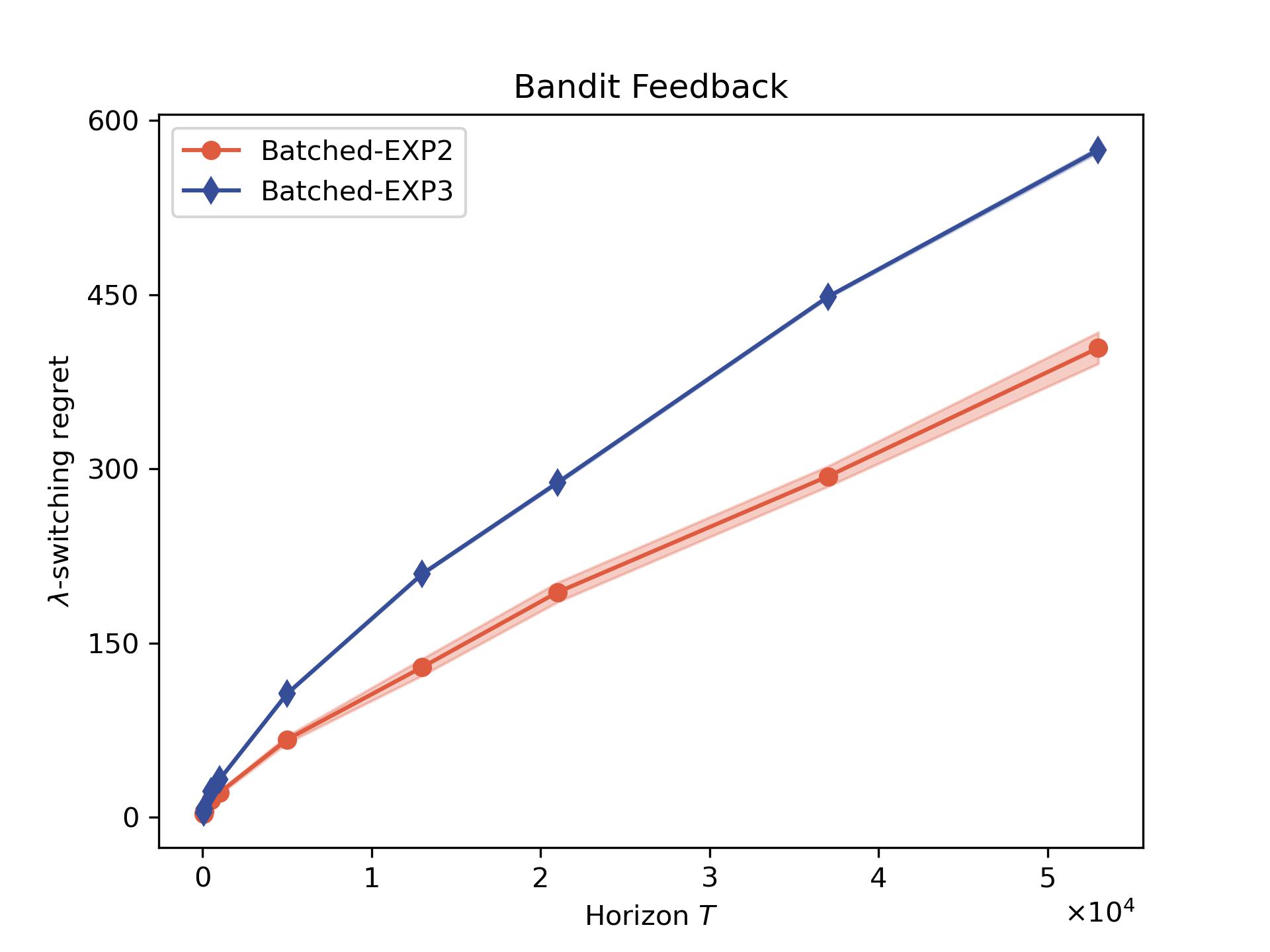}
		\caption{ $K=10$, $I=3$ and $\lambda = 0.1$ under  bandit feedback using the SC$(0.01)$ adversary. }
		\label{fig:banditcomparels0.1}
	\end{subfigure}
 \hfill
  \begin{subfigure}{0.32\textwidth}
		\includegraphics[width=\textwidth]{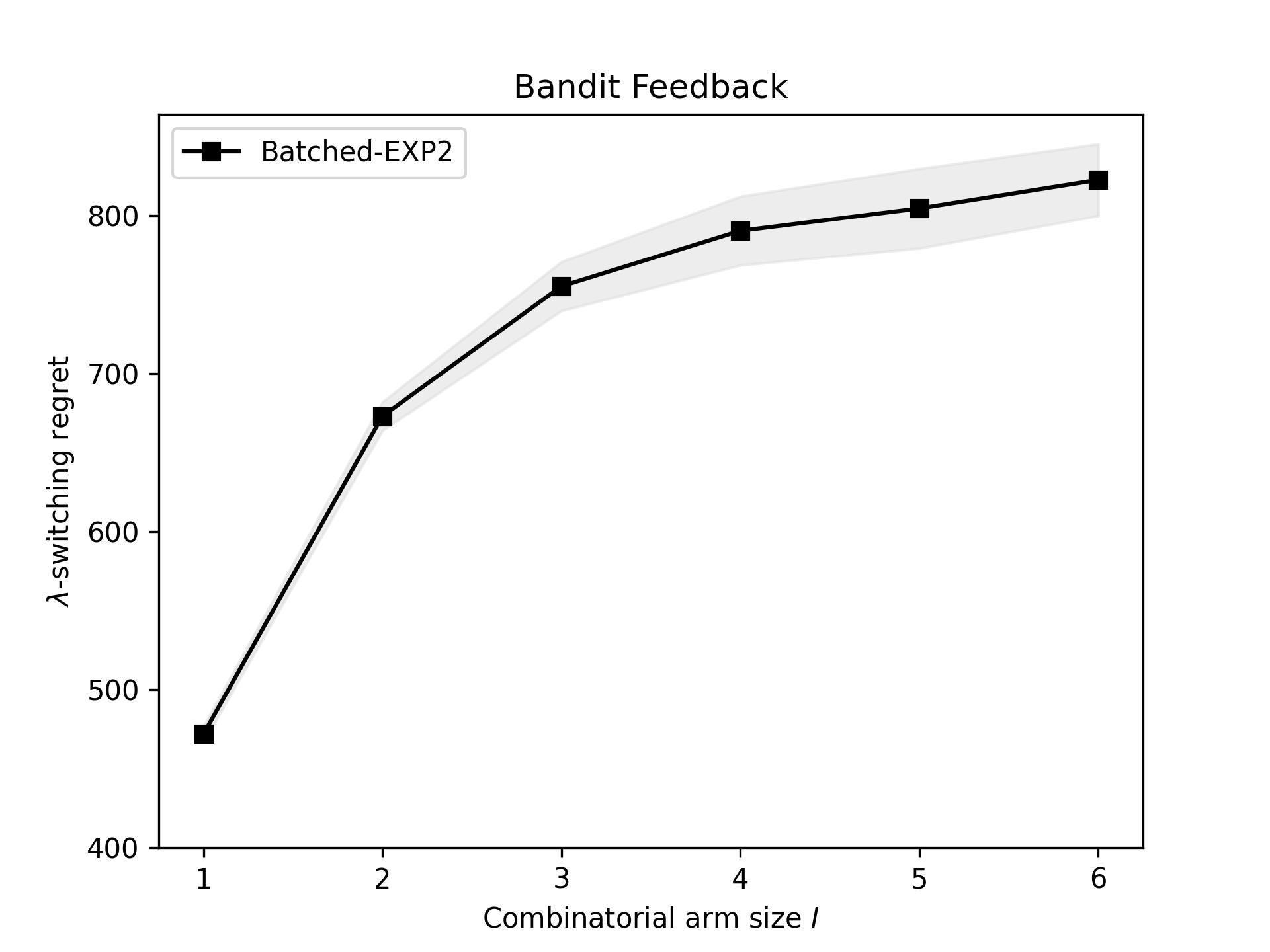}
		\caption{$K=20$, $T=10000$ and $\lambda = 1$ under bandit feedback using the CIN loss sequence.}
		\label{fig:figure2}
	\end{subfigure}
\hfill
\begin{subfigure}{0.32\textwidth}
		\includegraphics[width=\textwidth]{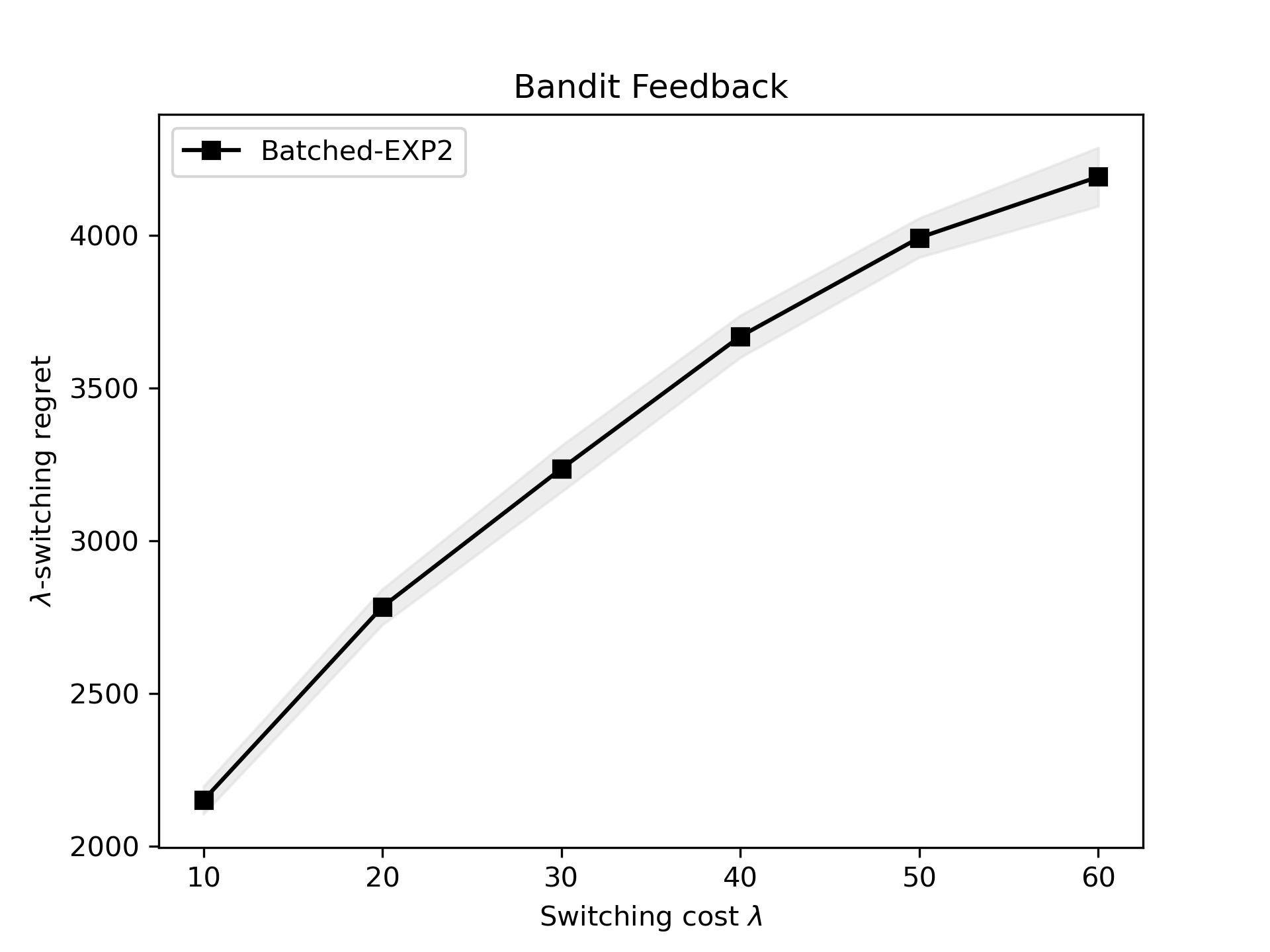}
		\caption{$K=30$, $T=10000$ and $I = 3$ under bandit feedback using the CIN loss sequence.}
		\label{fig:banditls}
	\end{subfigure}

	\caption{Comparison of the performance of different algorithms under Bandit Feedback}
	\label{fig:overall}
\end{figure*}


In this section, we present the numerical results to compare our algorithms \textsc{Batched-Exp2} with John's exploration in Algorithm~\ref{alg:EXP2John} and \textsc{Batchd-BROAD} in Algorithm~\ref{alg:BROAD} with some baselines from the literature after adding batches in which the played combinatorial arm does not change within each batch.  For bandit feedback, our baseline algorithm is the  \textsc{Exp$3$} algorithm~\cite{audibert2014regret} and we modified it to be a batched algorithm called \textsc{Batched-Exp$3$}.  For the semi-bandit feedback, we choose the Follow-the-Regularized-Leader  algorithm with  hybrid regularizer~\cite{zimmert2019beating} 
$
    F(a) = \sum_{i=1}^K -\sqrt{a_i} + \gamma (1-a_i)\log (1-a_i)
$
and the unnormalized negentropy potential~\cite{lattimore2020bandit} $F(a) = \sum_{i=1}^K \big(a_i\ln a_i-a_i\big)$. We call the modification of these two algorithms the \textsc{Batched-HYBRID} and \textsc{Batched-NegENTROPY}, respectively.

Since the optimal adversarial adversary is difficult to design, we use the CIN loss sequence and CDN loss sequence given by Algorithms~\ref{alg:sequence1} and~\ref{alg:sequence2} in~\S\ref{sec:lowerbound} for bandit and semi-bandit feedback, respectively.
\yan{Besides the lower-bound traces, we also design a stochastically constrained (SC) adversary which is very similar to that used in}~\cite{zimmert2019beating}. \yan{Specifically, the time horizon of length $T$ is split into phases:}
\begin{align*}
    \yan{\underbrace{1,2,\dots, t_1}_{T_1}, \underbrace{t_1+1,\dots,t_2}_{T_2}, \dots,\underbrace{t_{n-1},\dots, T}_{T_n},
    }
\end{align*}
\yan{where the length of phase $i$ is $T_i =\lfloor 1.6^i\rfloor $. The loss for each arm $i$ at  time $t$ is set to be an independent Bernoulli distribution with mean $$
    \mu_{ti}=\begin{cases}
        1-\Check{\alpha}\lambda&\text{ if }i\leq I\\
        1 & \text{else}
    \end{cases} $$ if $t $ belongs to an odd phase and
$$
     \mu_{ti}=\begin{cases}
        0&\text{ if }i\leq I\\
        \Check{\alpha}\lambda & \text{else}
    \end{cases} $$ otherwise. In the above setting, $\lambda $ is the switching cost. We denote the SC adversary with parameter $\Check{\alpha}$ by SC($\Check{\alpha}$) adversary. Note that the mean of the optimal arm oscillates between being close to 1 and close to 0 to create a challenging environment
    for our bandit algorithms.} 

\subsection{Bandit Feedback}\label{subsec:expbandit}
We use two types of adversaries to compare the performance of \textsc{Batched-Exp2} with John's exploration and  \textsc{Batched-Exp3} algorithm.
The batch length in the algorithms is fixed to be $B = \left \lceil{3\lambda^{\frac{2}{3}}K^{-\frac{1}{3}}(TI)^{\frac{1}{3}}}\right \rceil $.

First, we use the lower-bound trace CIN adversary that we designed in Algorithm~\ref{alg:sequence1} with $
   \sigma =\frac{10}{9 \log_2 T}$ and $  \epsilon = \frac{10(\lambda K)^{\frac{1}{3}}(IT)^{-\frac{1}{3}}}{9\log_2 T} .$ 
From Figure~\ref{fig:figure1}, we observe that the $\lambda$-switching regret of \textsc{Batched-Exp2} with John's exploration is much smaller than that of \textsc{Batched-Exp3} when $K=10$, $I=3$, $\lambda =1$. 
From Figure~\ref{fig:figure2a}, we observe that the $\lambda$-switching regret of \textsc{Batched-Exp2} with John's exploration is much smaller than that of \textsc{Batched-Exp3}  for a smaller value of $\lambda =0.1$, showing that even when the switching cost is small, our algorithm outperforms the benchmark significantly.
In Figure~\ref{fig:figure2}, we compare the $\lambda$-switching regret for different values of $I$ when $K=20$, $T=10000$ and $\lambda =1$.
It is observed that the regret grows as  $I^{0.304}$; the dependence appears to be loose with respect to the upper bound of $I^{\frac{4}{3}}$ in Theorem~\ref{thm:exp2} and suggests that the \textsc{Batched-Exp2} algorithm works well under the CIN loss sequence. Also, we observe that  $I^{0.304}$ is also loose in terms of lower bound $I^{\frac{2}{3}}$ in Theorem~\ref{thm:lowerbound}, which can be explained by the following statement. Under the CIN loss sequence $L_{1:T}$ and the policy $\pi^{\mathrm{Exp2}}$ of the \textsc{Batched-Exp2} algorithm, we delineate two reasons that explain why the $\lambda$-switching regret $R_{\lambda}(\pi^{\mathrm{Exp2}}, L_{1:T})$ is not lower bounded by $\Omega(I^{\frac{2}{3}})$.
   \begin{figure*}[t]
	\centering
	\begin{subfigure}{0.32\textwidth}
		\includegraphics[width=\textwidth]{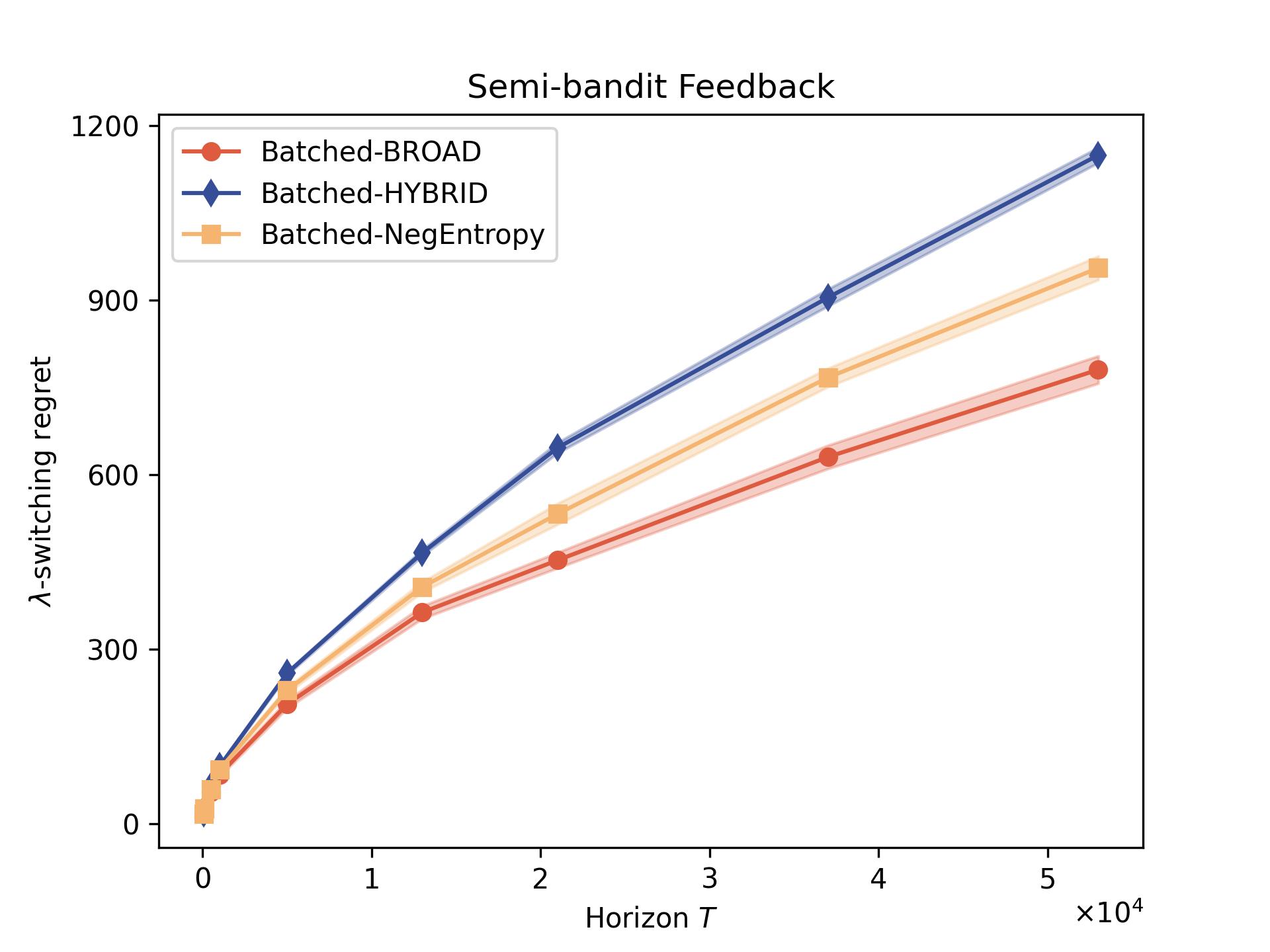}
		\caption{ $K=10$, $I=3$ and $\lambda = 1$ under semi-bandit feedback using the CDN loss sequence. }
		\label{fig:semi1}
	\end{subfigure}
	\hfill
  	\begin{subfigure}{0.32\textwidth}
		\includegraphics[width=\textwidth]{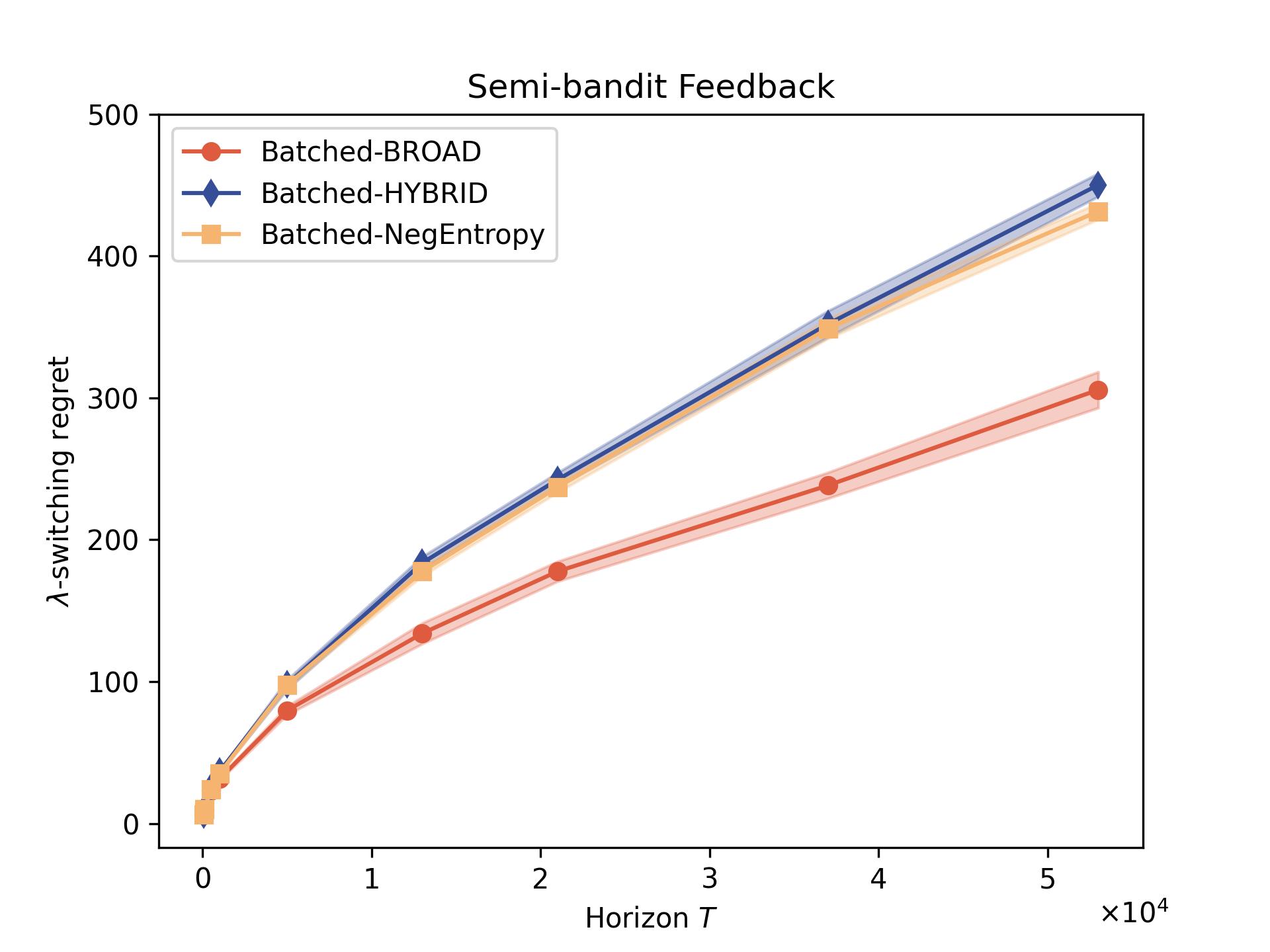}
		\caption{ $K=10$, $I=3$ and $\lambda = 0.1$ under semi-bandit feedback using the CDN loss sequence. }
		\label{fig:figure2b}
	\end{subfigure}
 \hfill
	\begin{subfigure}{0.32\textwidth}
		\includegraphics[width=\textwidth]{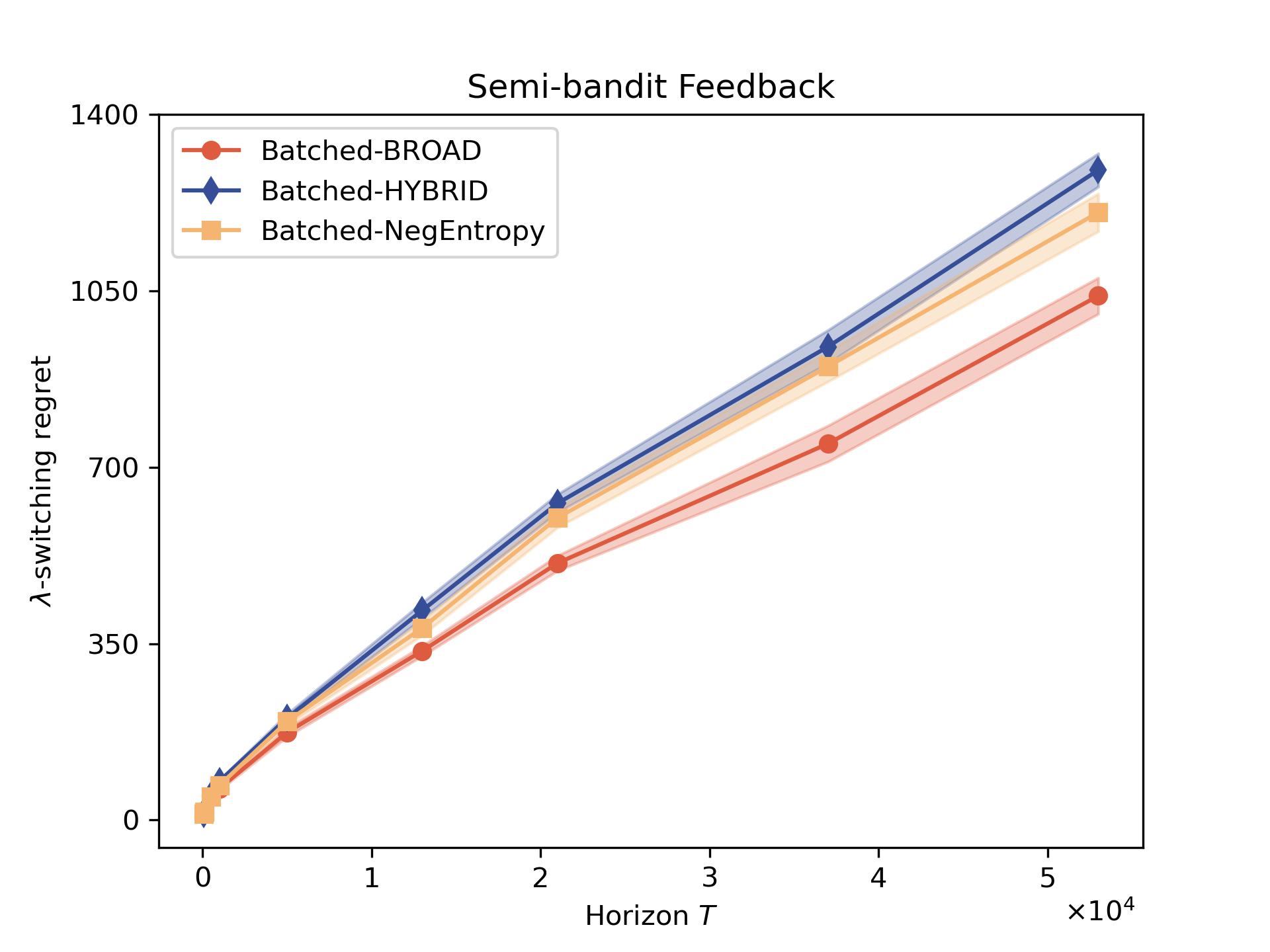}
		\caption{ $K=10$, $I=3$ and $\lambda = 1$ under semi-bandit feedback using the SC$(0.005)$ adversary. }
		\label{fig:semi1_v2}
	\end{subfigure}
	\hfill
  	\begin{subfigure}{0.32\textwidth}
		\includegraphics[width=\textwidth]{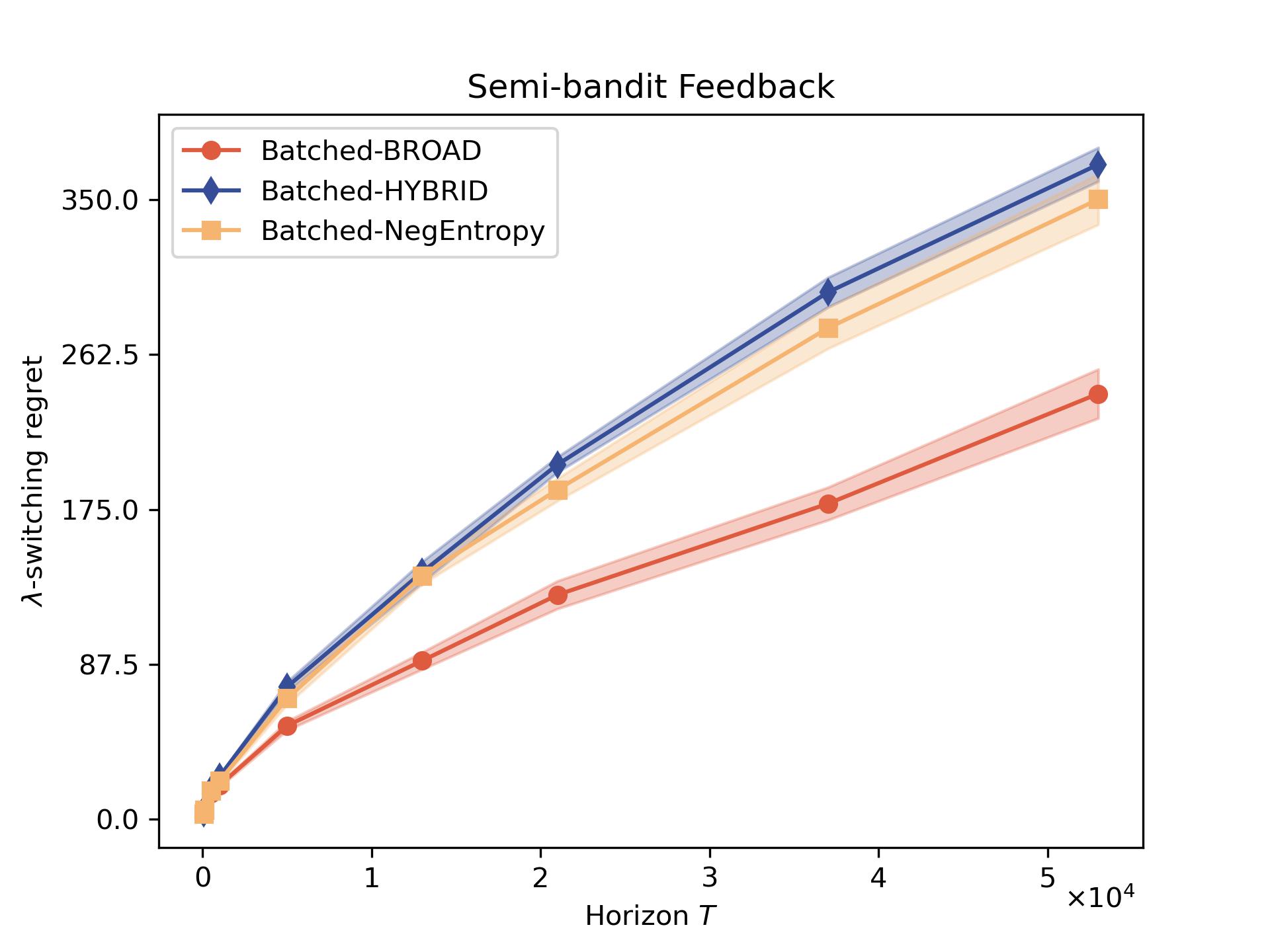}
		\caption{ $K=10$, $I=3$ and $\lambda = 0.1$ under semi-bandit feedback using the SC$(0.005)$ adversary. }
		\label{fig:figure2b_v2}
	\end{subfigure}
 \hfill
 \begin{subfigure}{0.32\textwidth}
		\includegraphics[width=\textwidth]{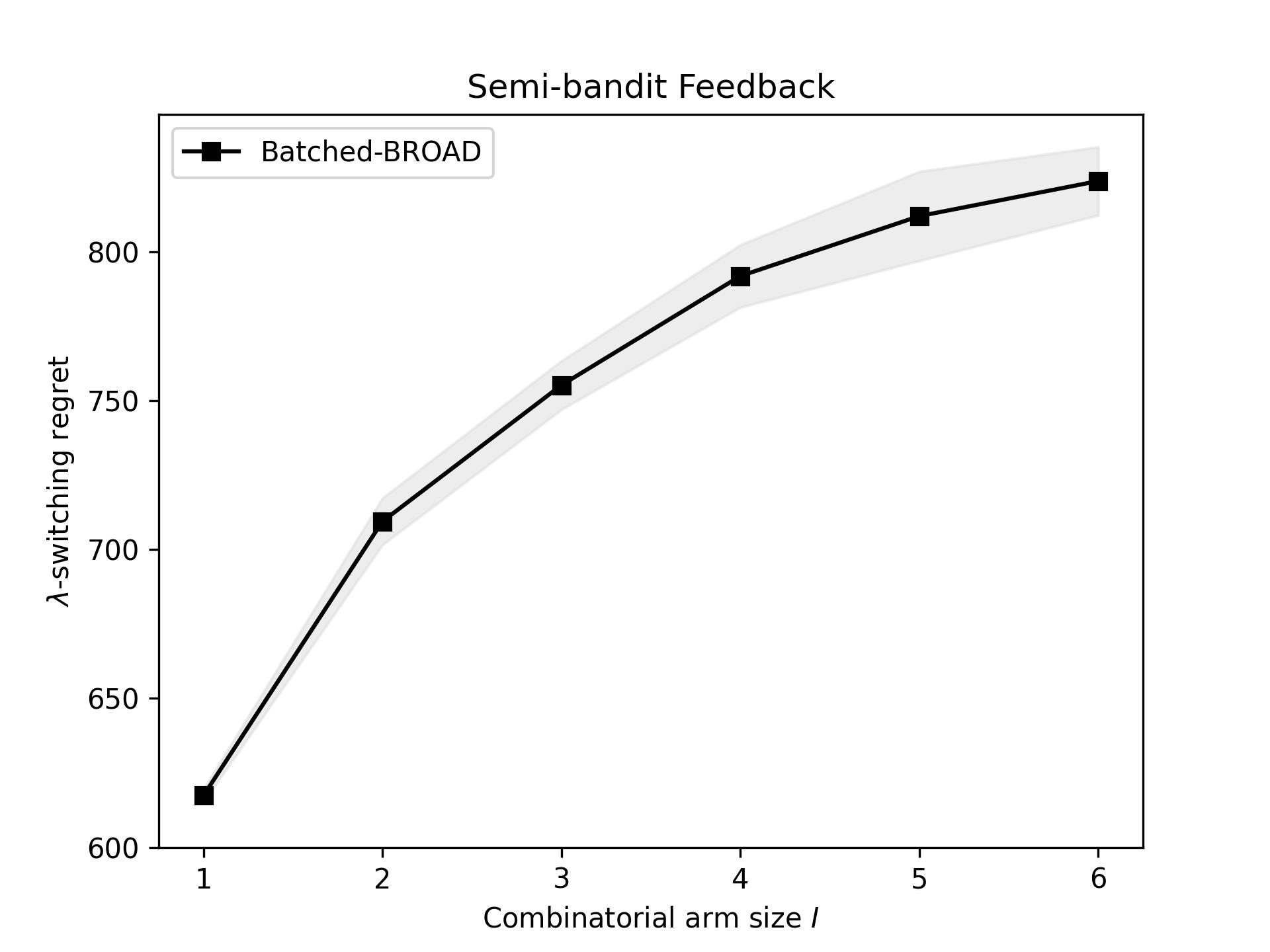}
		\caption{$K=40$, $T=10000$ and $\lambda = 1$ under semi-bandit feedback using the CDN loss sequence.}
		\label{fig:semi2}
	\end{subfigure}
 \hfill
	\begin{subfigure}{0.32\textwidth}
		\includegraphics[width=\textwidth]{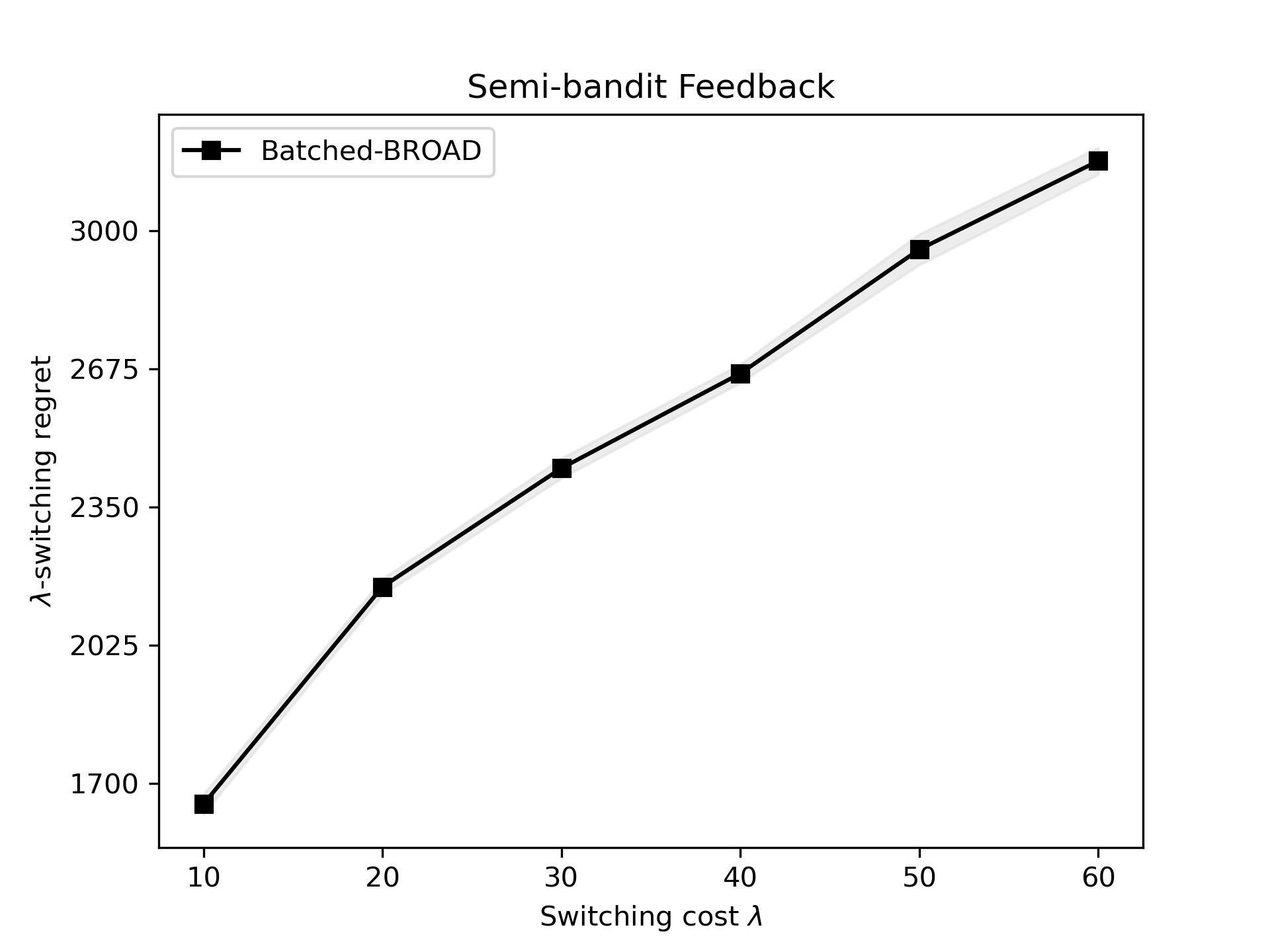}
		\caption{$K=30$, $T=10000$ and $I = 3$ under semi-bandit feedback using the CDN loss sequence.}
		\label{fig:semi2_v2}
	\end{subfigure}
	\caption{Comparison of the performances of different algorithms  under  Semi-bandit Feedback}
	\label{fig:overall_semi}
\end{figure*}


 

\begin{enumerate}
    \item Let $\mc V$ be the set of all stochastic loss sequences $V_{1:T}$ ($V_t\in [0,1]^K$ is a random vector). Given the combinatorial arm size $K$, the switching cost $\lambda$ and the time horizon $T$, we have 
    \begin{align*}
        &\inf_{\pi \in \Pi} \sup_{V_{1:T}\in \mc V} R_{\lambda} (\pi, V_{1:T}) \nonumber\\*
        &\qquad \geq    \inf_{\pi \in \Pi} \sup_{l_{1:T}\in \mc L}   R_{\lambda} (\pi, l_{1:T}) \geq \Omega(I^{\frac{2}{3}}),
    \end{align*}
    where the last inequality holds due to Theorem~\ref{thm:lowerbound}. It is not comparable between $R_{\lambda}(\pi^{\mathrm{Exp2}}, L_{1:T}) $ and $  \inf_{\pi \in \Pi} \sup_{V_{1:T}\in \mc V} R_{\lambda} (\pi, V_{1:T}).$ Thus $R_{\lambda}(\pi^{\mathrm{Exp2}}, L_{1:T})$ may not be $\Omega(I^{\frac{2}{3}})$.
    \item Given the combinatorial arm size $K$, the switching cost $\lambda$ and the time horizon $T$, by Lemma~\ref{lm:lowerbound1},  for any deterministic player $\pi$, we have 
    \begin{equation*}
          R_{\lambda} (\pi, L_{1:T}) \geq \Omega(I^{\frac{2}{3}}),
    \end{equation*}
     under the CIN loss sequence $L_{1:T}$. Since the policy $\pi^{\mathrm{Exp2}}$ of the \textsc{Batched-Exp2} algorithm is stochastic, $R_{\lambda}(\pi^{\mathrm{Exp2}}, L_{1:T})$ may not be $\Omega(I^{\frac{2}{3}})$.
\end{enumerate}
\yan{In Figure~\ref{fig:banditls}, we compare the $\lambda$-switching regret for different values of $\lambda$ when $K=30$, $T=10000$, and $I=3$. It is observed that the regret grows as $\lambda^{0.379}$.
Note that our theoretical results in Theorems~\ref{thm:lowerbound} and~\ref{thm:exp2} say that the expected $\lambda$-switching regret scales as $\Theta(\lambda^{1/3})$ when $T$, $K$ and $I$ are fixed. Even though the empirical observation of the regret scaling as $\lambda^{0.379}$ cannot be directly compared to the theoretical result of $\lambda^{1/3}$ because, among other reasons, the loss sequence constructed here is, in fact, stochastically constrained, the fact that the exponents of $\lambda$ are not too far from each other is reassuring.  
}

\yan{The second trace we used is the SC$(0.01)$ adversary. From Figure~\ref{fig:banditcomparels1.0} and Figure~\ref{fig:banditcomparels0.1}, we observe that the $\lambda$-switching regrets of \textsc{Batched-Exp2} with John's exploration are both smaller than those of \textsc{Batched-Exp3} when $K=10$, $I=3$ and $\lambda =1$, and $K=10$, $I=3$ and $\lambda =0.1$, respectively. This again corroborates the efficacy of our proposed methods. }

\subsection{Semi-bandit Feedback}
We compare \textsc{Batched-BROAD}, \textsc{Batched-HYBRID} and  \textsc{Batched-NegENTROPY} algorithm under the lower-bound trace CDN adversary that we designed in Algorithm~\ref{alg:sequence2} with 
$ \sigma =\frac{10}{ 9 \log_2 T } $ and $ \epsilon = \frac{10(\lambda k)^{\frac{1}{3}}I^{-\frac{2}{3}}T^{-\frac{1}{3}}}{9\log_2 T} $ and the SC$(0.005)$ adversary. The batch length of the algorithms is fixed to be $B = \left \lceil{3\lambda^{\frac{2}{3}}K^{-\frac{1}{3}}T^{\frac{1}{3}}I^{\frac{2}{3}}}\right \rceil $.

Under the CDN adversary, we have the following results. From Figure~\ref{fig:semi1}, we observe that the $\lambda$-switching regret of \textsc{Batched-BROAD} is much smaller than that of \textsc{Batched-HYBRID} and \textsc{Batched-NegENTROPY} when $K=10$, $I=3$, $\lambda =1$. From Figure~\ref{fig:figure2b}, we observe that the $\lambda$-switching regret of \textsc{Batched-BROAD} is much smaller than that of \textsc{Batched-HYBRID} and \textsc{Batched-NegENTROPY} for a smaller value $\lambda =0.1$, showing that even when the switching cost is small, our algorithm outperforms the benchmark significantly. In Figure~\ref{fig:semi2}, we compare the $\lambda$-switching regret for different values of $I$ when $K=40$, $T=10000$ and $\lambda =1$. 
It is observed that the regret grows as  $I^{0.163}$; the dependence appears to be loose with respect to the upper bound of $I^{\frac{2}{3}}$ in Theorem~\ref{thm:broad} and suggests that the \textsc{Batched-BROAD} algorithm works well under the CDN loss sequence. Also, we observe that  $I^{0.163}$ is also loose in terms of lower bound $I^{\frac{1}{3}}$ in Theorem~\ref{thm:lowerbound-semi}, which is possible by a  similar reasoning as that for bandit feedback in~\S\ref{subsec:expbandit}.
\yan{In Figure~\ref{fig:semi2_v2}, we compare the $\lambda$-switching regret for different values of $\lambda$ when $K=30$, $T=10000$, and $I=3$. It is observed that the regret grows as $\lambda^{0.356}$, which is again close to $\lambda^{1/3} $ as given by our theoretical result in Theorems~\ref{thm:lowerbound-semi} and~\ref{thm:broad}.} 

\yan{Under the SC$(0.005)$ adversary, from Figure~\ref{fig:semi1_v2} and Figure~\ref{fig:figure2b_v2}, we observe that the $\lambda$-switching regrets of \textsc{Batched-BROAD} are both smaller than that of \textsc{Batched-HYBRID} and \textsc{Batched-NegENTROPY}  when $K=10$, $I=3$ and $\lambda =1$, and $K=10$, $I=3$ and $\lambda =0.1$, respectively.}
 \section{Conclusion}
 We derived lower bounds for the minimax regret for the problem of adversarial combinatorial bandit with a switching cost $\lambda$ for each changed arm in each round. We also designed algorithms that operate in batches to approach the lower bounds. Our findings provide insights into the inherent difficulty of the problem and suggest efficient approaches to minimize switching costs and optimize the overall performance in terms of regret. Further research involves deriving tighter bounds in both directions for both bandit and semi-bandit feedback. Also, other sets of combinatorial arms such as those involved in the shortest path problem, ranking problems, and multitask problems can also be considered when switching costs are involved.

		\appendix
		\subsection{Proof of Lemma~\ref{lm:lowerbound1}}\label{app:banditlower}
		Given the constructed stochastic loss sequence $L_{1:T}$ defined in Algorithm~\ref{alg:sequence1}, we now want to analyze the player's expected regret under an arbitrary  deterministic policy $\pi$ which  yields an action sequence $ A_{1:T} \in \mc A^T$ so that $A_t$ is a function of the player's past observations $X_{1:t-1} $ with $X_t  = \langle A_t, L_t\rangle $.
  First we define
	\begin{align}		R\triangleq 	\sum_{t=1}^T   \langle A_t,L_t \rangle+   \lambda\sum_{t=1}^T   d(A_t,A_{t-1}) - \min_{A\in \mc A} \sum_{t=1}^T  \langle A,  L_t \rangle .\label{eq:Rdef_v2}
		\end{align}
		\yan{We also define the regret with respect to the unclipped stochastic loss functions $\tilde L_{1:T} $ defined in Algorithm~\ref{alg:sequence1}  under the same deterministic policy $\pi$ which  yields an action sequence $ \tilde A_{1:T} \in \mc A^T$ so that $\tilde A_t$ is a function of the player's past observations $Y_{1:t-1} $ with $Y_t  = \langle \tilde A_t, \tilde L_t\rangle $.} \yan{Let}
		\begin{align*}
			\yan{\tilde R \triangleq 	\sum_{t=1}^T  \langle \tilde A_t,\tilde{L}_t\rangle+  \lambda \sum_{t=1}^T   d(\tilde A_t,\tilde A_{t-1}) - \min_{A\in \mc A} \sum_{t=1}^T   \langle A
   , \tilde{L}_t  \rangle.}
		\end{align*}
		Then $\E[R]=R_{\lambda}(\pi, L_{1:T}) $ where the expectation in $\E[R]$ is taken over the adversary’s randomized choice of the loss sequences $L_{1:T}$, and $R_{\lambda}(\pi, L_{1:T})$ is defined in~\eqref{eq:defRlambdasto}.  
  	The next lemma shows that in expectation, the regret $\E[R]$ can be lower bounded in terms of $\E[\tilde R]$ (the expectation in $\E[\tilde R]$ is taken over the adversary’s randomized choice of the loss sequences $\tilde L_{1:T}$).
			
  	\begin{lemma}\label{lm:RRprime_v2}
			Assume that $ T\geq \max \{\frac{\lambda K}{I},6\} $. Then $ 	\E[R]\geq 	\E[\tilde R] -\frac{\epsilon TI}{4}. $
		\end{lemma}
		\begin{IEEEproof}
			We consider the event $ B=\{\forall\, t: L_t=\tilde{L}_t \} $, and first show that $ P(B)\geq 1-\frac{\epsilon}{4(\lambda+\epsilon)} $. For $ \delta = \frac{\epsilon}{4(\lambda+\epsilon)} $, by~\cite[Lemma 1]{dekel2014bandits} we have that with probability at least $ 1-\delta $, 
			\begin{align*}
				|W_t|\leq\sigma \sqrt{2d(\rho)\ln \frac{T}{\delta}} \leq 2\sigma \sqrt{ \log_2 T \log_2 \frac{4T(\lambda+\epsilon)}{\epsilon}},
			\end{align*}
   for all $t\in [T]$, where the last inequality holds due to $d(\rho)\leq \log_2 T+1$ by~\cite[Lemma 2]{dekel2014bandits}. Thus, setting $ \sigma =\frac{1}{6 \sqrt{ \log_2 T \log_2 \frac{4T(\lambda+\epsilon)}{\epsilon}}} $, we obtain that 
			\begin{align*}
				P\Big( \forall\, t\in [T] , \frac{1}{2} +W_t\in \Big[\frac{1}{6},\frac{5}{6}\Big]\Big)\geq 1-\delta.
			\end{align*}
			For $ T\geq \max(\frac{\lambda K}{I},6) $, we have $ \epsilon < \frac{1}{6} $ and thus $ \tilde{L}_t(x)\in [0,1] $ for all $ x\in [K] $ whenever $ 1/2+W_t \in [\frac{1}{6},\frac{5}{6}] $. This implies that $ P(B)\geq 1-\delta. $
			
			If $ B $ occurs then $ R=\tilde R $; otherwise, $ \tilde R-R\leq (\lambda+\epsilon) TI $ since $R, \tilde R\in [0, (\lambda + \epsilon)TI]$. Therefore,
			\begin{align*}
				\E[\tilde R]-\E[R] = \E[\tilde R-R| \lnot B]\cdot P(\lnot B)\leq \frac{\epsilon TI}{4}.
			\end{align*}
		\end{IEEEproof}
  
		
  Let $\mc Q_0$ and $\mc Q_{\mc I}$ follow previous definition in~\eqref{eq:Qdef1} and~\eqref{eq:Qdef2}, $ \tilde{\mathcal F} $
		be the $ \sigma $-algebra generated by $X_{1:T} $ defined in~\eqref{eq:ytsequence_v2}  and $d^{\tilde{\mathcal F}}_{\mathrm{TV}} (\mathcal Q_0, \mathcal Q_{\mathcal I}) $ follow the definition in~\eqref{eq:defdev_v3}.
  $\tilde M_{\mc I}$ and $\tilde M$ are defined in~\eqref{eq:defMI} and~\eqref{eq:defM}, respectively. Then we have the following lemma that bounds total variation from above.
		
	\begin{remark}\label{rm:algebra_v2}
	    Note that $\tilde A_t$ is a deterministic function of its past observations $Y_{1:t-1}$; thus the $\sigma$-algebra generated by $\tilde A_{1:T}$ is a subset of $\tilde{\mc F}.$
	\end{remark}	

		\begin{lemma}\label{cor:inequa}
			It holds that
			$$\yan{\frac{1}{\binom{K}{I}}  \sum_{\mathcal I \in \mc A} d^{\tilde{\mathcal F}}_{\mathrm{TV}} (\mathcal Q_0, \mathcal Q_{\mathcal I}) \leq  \frac{\epsilon }{\sigma \sqrt{2K}} \sqrt{(\log_2 T+1) E_{\mathcal Q_0}[\tilde M]}.}$$
		\end{lemma}
		\begin{IEEEproof}
   By~\cite[Lemma 2]{dekel2014bandits}, the width $w(\rho)\leq \log_2 T+1$.
			 Then by Lemma~\ref{lm:partswitch1}, 
    we have 
			\begin{align*}
				d^{\tilde{\mathcal F}}_{\mathrm{TV}} (\mathcal Q_0, \mathcal Q_{\mathcal I}) \leq \frac{\epsilon}{2\sigma\sqrt{I}} \sqrt{(\log_2 T+1)  \E_{\mathcal Q_0}[\tilde M_{\mathcal I}]}.
			\end{align*}
   Then using the concavity of the squared root function and by $ \sum_{\mathcal I \in \mc A} \tilde{M}_{\mathcal I} = 2\binom{K-1}{I-1} \tilde M$, it holds that
			\begin{align*}
				&\frac{1}{\binom{K}{I}}  \sum_{\mathcal I \in \mc A}	d^{\tilde{\mathcal F}}_{\mathrm{TV}} (\mathcal Q_0, \mathcal Q_{\mathcal I})  \nonumber\\*
    &\leq \frac{\epsilon}{2\sigma\sqrt{I}} \sqrt{ \log_2 T+1 }  \sum_{\mathcal I \in \mc A} \frac{1}{\binom{K}{I}} \sqrt{ \E_{\mathcal Q_0}[\tilde M_{\mathcal I}]} \\
				&\leq \frac{\epsilon}{2\sigma\sqrt{I}} \sqrt{\log_2 T+1  } \sqrt{ \E_{\mathcal Q_0}\Bigg[ \frac{1}{\binom{K}{I}}  \sum_{\mathcal I \in \mc A}\tilde M_{\mathcal I}\Bigg]}\\
				&= \frac{\epsilon}{\sqrt{2}\sigma} \sqrt{\frac{ (\log_2 T+1) \E_{\mathcal Q_0}[  \tilde M] }{K}}.
			\end{align*}
		\end{IEEEproof}

		\begin{lemma}\label{lm:Rprime}
  It holds that
			\begin{align*}
				\yan{\E[\tilde R] \geq \epsilon TI \Big(1-\frac{I}{K}\Big) - \frac{\epsilon  TI}{\binom{K}{I}} \sum_{\mathcal I\in \mc A} d^{\tilde{\mathcal F}}_{\mathrm{TV}} (\mathcal Q_0,\mathcal Q_{\mathcal I}) + \lambda\E[\tilde M].}
			\end{align*}
		\end{lemma}
		\begin{IEEEproof}
			For any $ i\in [K]$, let $ T_{ i} $ denote the number of rounds the player picks arm $i$  in the action sequence $\tilde A_{1:T}$.
			So we can write $ \tilde R  = \epsilon \Big(TI -\sum_{i\in [K]} \chi_i T_i\Big) +\lambda \tilde M $, where we use $\chi_i$ to denote the $i$-th component of $\chi$. Also,  we use $\mc I_i$ to denote the $i$-th component of $\mc I\in \mc A$ in the following. Since $\chi \in \mc A$ is selected uniformly at random in Algorithm~\ref{alg:sequence1}, we have
			\begin{align*}
				\E[\tilde R] &= \frac{1}{\binom{K}{I}} \sum_{\mc I\in \mc A} \E\Big[   \epsilon \Big(TI -\sum_{i\in [K]} \mc I_i T_i\Big) +\lambda \tilde M  \Big | \chi=\mc I \Big]\\
				&= \epsilon TI - \frac{\epsilon}{\binom{K}{I}} \sum_{\mc I\in \mc A} \sum_{i\in [K]} \mc I_i\E_{Q_{\mc I}}\big[T_i\big] +\lambda\E[\tilde M].
			\end{align*}
			For all $ i\in [K]$ and $ t\in [T] $, the event $\{ \tilde A_{t,i}=1 \} $ belongs to the $ \sigma $-field $ \tilde {\mathcal F} $ by Remark~\ref{rm:algebra_v2} ($\tilde A_{x,i}$ is used to denote the $i$-th component of $\tilde A_x$), so we have
			\begin{align*}
				\mathcal Q_{\mathcal I}(\tilde A_{t,i}=1 ) - \mathcal Q_0(\tilde A_{t,i}=1) \leq d^{\tilde{\mathcal F}}_{\mathrm{TV}} (\mathcal Q_0,\mathcal Q_{\mathcal I}).
			\end{align*}
			Summing over $ t\in[T] $ yields
			\begin{align*}
				\E_{\mathcal Q_{\mathcal I}} \big[T_i\big] - 	\E_{\mathcal Q_{0}} \big[T_i\big] \leq Td^{\tilde{\mathcal F}}_{\mathrm{TV}} (\mathcal Q_0,\mathcal Q_{\mathcal I}).
			\end{align*}
   Summing over $i\in [K]$ such that $ \mc I_i=1 $ yields
			\begin{align*}
				\sum_{i\in [K]}	\mc I_i \Big(	\E_{\mathcal Q_{\mathcal I}} \big[T_i\big] - 	\E_{\mathcal Q_{0}} \big[T_i\big] \Big)\leq TId^{\tilde{\mathcal F}}_{\mathrm{TV}} (\mathcal Q_0,\mathcal Q_{\mathcal I}).
			\end{align*}
   Summing over $ \mc I\in\mc A $ yields
			\begin{align*}
				\sum_{\mathcal I\in \mc A}\sum_{i\in [K]} \mc I_i\Big(	\E_{\mathcal Q_{\mathcal I}} \big[T_i\big] - 	\E_{\mathcal Q_{0}} \big[T_i\big]\Big)\leq TI	\sum_{\mathcal I\in \mc A} d^{\tilde{\mathcal F}}_{\mathrm{TV}} (\mathcal Q_0,\mathcal Q_{\mathcal I}).
			\end{align*}
			Thus
			\begin{align*}
			&	\sum_{\mathcal I\in \mc A}\sum_{i\in [K]}\mc I_i\E_{\mathcal Q_{\mathcal I}}  \big[T_i\big] \nonumber\\*
    &\leq 	\sum_{\mathcal I\in \mc A}\sum_{i\in [K]}	\mc I_i \E_{\mathcal Q_{0}}  \big[T_i\big]  + TI	\sum_{\mathcal I\in \mc A} d^{\tilde{\mathcal F}}_{\mathrm{TV}} (\mathcal Q_0,\mathcal Q_{\mathcal I}) \\
				& =\binom{K-1}{I-1} TI+TI	\sum_{\mathcal I\in \mc A} d^{\tilde{\mathcal F}}_{\mathrm{TV}} (\mathcal Q_0,\mathcal Q_{\mathcal I}).
			\end{align*}
			Therefore,
			\begin{align*}
				\E[\tilde R] &\geq \epsilon TI - \frac{\epsilon}{\binom{K}{I}}\Big( \binom{K-1}{I-1} TI+TI	\sum_{\mathcal I\in \mc A} d^{\tilde{\mathcal F}}_{\mathrm{TV}} (\mathcal Q_0,\mathcal Q_{\mathcal I})\Big) \nonumber\\*
    &\qquad+\lambda \E[\tilde M]\\
				&=\epsilon TI \Big(1-\frac{I}{K}\Big) - \frac{\epsilon  TI}{\binom{K}{I}} \sum_{\mathcal I\in \mc A} d^{\tilde{\mathcal F}}_{\mathrm{TV}} (\mathcal Q_0,\mathcal Q_{\mathcal I}) +\lambda \E[\tilde M].
			\end{align*}
			
		\end{IEEEproof}

		\begin{IEEEproof}[Proof of Lemma~\ref{lm:lowerbound1}]
			We first prove Lemma~\ref{lm:lowerbound1} for deterministic policies that make no more than $S_0= \frac{\epsilon T I}{\lambda}$ switches. For algorithms with this property, we have 
			\begin{align*}
				\mathcal Q_0(\tilde M>  \epsilon T I)=	\mathcal Q_{\mathcal I}(\tilde M>  \epsilon T I) =0.
			\end{align*}
   As the event $\{\tilde M\geq m\}$ is in $\tilde {\mc F}$ by Remark~\ref{rm:algebra_v2}, then 
			\begin{align*}
				\E_{\mathcal Q_0}[\tilde M] - \E_{\mathcal Q_{\mathcal I}}[\tilde M]  &= \sum_{m=1}^{S_0} \Big(\mathcal Q_0 (\tilde M\geq m)- \mathcal Q_{\mathcal I} (\tilde M\geq m) \Big) \\
				&\leq \frac{\epsilon TI}{\lambda} d^{\tilde{\mathcal F}}_{\mathrm{TV}} (\mathcal Q_0, \mathcal Q_{\mathcal I}).
			\end{align*}
			Then 
			\begin{align}
				\E_{\mathcal Q_0}[\tilde M] - \E[\tilde M]   &= \frac{1}{\binom{K}{I}} \sum_{\mathcal I \in \mc A}
				\Big( \E_{\mathcal Q_0}[\tilde M]- \E_{\mathcal Q_{\mathcal I}}[\tilde M] \Big)\nonumber\\
				&\leq \frac{ \epsilon TI}{\lambda\binom{K}{I}} \sum_{\mathcal I \in \mc A}d^{\tilde{\mathcal F}}_{\mathrm{TV}} (\mathcal Q_0, \mathcal Q_{\mathcal I}).\label{eq:minequa}
			\end{align}
			Combining~\eqref{eq:minequa} with Lemma~\ref{lm:RRprime_v2} and Lemma~\ref{lm:Rprime}, we obtain
   			\begin{align*}
			\E[R]&\geq \epsilon TI \Big(1-\frac{I}{K} -\frac{ 1}{4 }\Big) -  \frac{2\epsilon  TI}{\binom{K}{I}} \sum_{\mathcal I\in \mc A} d^{\mathcal F}_{\mathrm{TV}} (\mathcal Q_0,\mathcal Q_{\mathcal I})  \nonumber\\*
   &\qquad + \lambda \E_{\mathcal Q_0} [\tilde M].
			\end{align*}
			By Lemma~\ref{cor:inequa}, and $ \log_2 T+1\leq 2\log_2 T $, we have
			\begin{align*}
				\frac{1}{\binom{K}{I}}  \sum_{\mathcal I \subset [K]} d^{\tilde{\mathcal F}}_{\mathrm{TV}} (\mathcal Q_0, \mathcal Q_{\mathcal I}) 
				&\leq   \frac{\epsilon  }{\sigma \sqrt{K}} \sqrt{(\log_2 T) \E_{\mathcal Q_0}[\tilde M]}.
			\end{align*}
			\yan{Using the notation $ m= \sqrt{\E_{\mathcal Q_0}[\tilde M]} $ and when $ K\geq 3 I $,}
   			\begin{align*}
				\yan{\E[R] \geq \frac{5}{12}\epsilon TI - \frac{2\epsilon^2  TI}{\sigma \sqrt{K}} \sqrt{\log_2 T }m +\lambda m^2,}
			\end{align*}
			\yan{where the right hand side is minimized when $ m = \frac{\epsilon^2 TI \sqrt{\log_2 T}}{\lambda \sigma \sqrt{K}} $. Thus the right-hand side is lower bounded by 
   $   \frac{5\epsilon TI}{12} - \frac{\epsilon^4 T^2I^2 \log_2 T}{\lambda\sigma^2 K}$}. \yan{Using our choice of 
   $ \sigma =\frac{1}{6 \sqrt{ \log_2 T \log_2 \frac{4T(\lambda+\epsilon)}{\epsilon}}} $ and $ \epsilon = \frac{(\lambda K)^{\frac{1}{3}}(IT)^{-\frac{1}{3}}}{9\log_2 T} $, we derive}
						\begin{align}
				\yan{\E[R]\geq \frac{5(\lambda K)^{\frac{1}{3}} (TI)^{\frac{2}{3}}}{108\log_2 T}-\frac{4(\lambda K)^{\frac{1}{3}} (TI)^{\frac{2}{3}} \log_2(\frac{4T(\lambda+\epsilon)}{\epsilon})}{9^3(\log_2 T)^2}.\label{eq:regret_v2}}
			\end{align}
   \yan{If $\lambda <\epsilon$, the right hand side of~\eqref{eq:regret_v2} is lower bounded by $\frac{(\lambda K)^{\frac{1}{3}} (TI)^{\frac{2}{3}}}{30\log_2 T}$ under the assumption that $T\geq 8$. Otherwise, if $\epsilon\leq \lambda \leq T$, we have
   }
   \begin{align}
       \yan{\epsilon = \frac{(\lambda K)^{\frac{1}{3}}(IT)^{-\frac{1}{3}}}{9\log_2 T} \geq \frac{(3\lambda)^{1/3}}{9 (\log_2 T) T^{1/3}} \geq \frac{(\lambda)^{1/3}}{7  T^{4/3}}, }\label{eq:inequaEps}
   \end{align}
   \yan{where the first inequality is due to $K\geq 3 I$. Combining $\lambda \geq \epsilon$ and \eqref{eq:inequaEps}, we get $\lambda \geq \frac{T^{-2}}{20}$ and then $\epsilon \geq \frac{2T^{-2}}{39}$. Thus when $\epsilon \leq \lambda \leq T$ and $T\geq 8$, the right hand side of~\eqref{eq:regret_v2} is lower bounded by $\frac{(\lambda K)^{\frac{1}{3}} (TI)^{\frac{2}{3}}}{130\log_2 T}$. Therefore, for any $K\geq 3I$ and $T\geq \max \{\frac{\lambda K}{I},8\}$, it holds that 
   }
   \begin{align}
       \yan{\E[R]\geq \frac{(\lambda K)^{\frac{1}{3}} (TI)^{\frac{2}{3}}}{130\log_2 T}} \label{eq:lower_bound_v2}
   \end{align}

			For any general algorithm that has an arbitrary number of switches, we can turn it to a new algorithm that makes at most $S_0$ switches by halting the algorithm once it makes $S_0$ switches and repeating the last action in the remaining rounds. The regret $R^*$ of the new algorithm (as defined in~\eqref{eq:Rdef_v2} under new algorithm) equals $R$ when $M\leq S_0$ and when $M> S_0$,
			\begin{align*}
				R^* \leq R+\epsilon TI \leq 2R,
			\end{align*}
			since $R\geq \lambda S_0$. Thus $\E[R^*]\leq 2\E[R]$. Since $\E[R^*]$ is lower bounded by the right-hand side of \eqref{eq:lower_bound_v2}, this implies the claimed lower bound on the expected regret of any deterministic player.
		\end{IEEEproof}
		
		\subsection{Proof of Lemma~\ref{lm:lowerbound2}}\label{app:semibanditlower}
  		Given the constructed stochastic loss sequence $L_{1:T}$ defined in Algorithm~\ref{alg:sequence2}, we now want to analyze the player's expected regret under an arbitrary deterministic policy $\pi$ which yields an action sequence $ A_{1:T} \in \mc A^T$ so that $A_t$ is a function of the player's past observations $Z_{1:t-1} $ with $Z_t  =  A_t\circ L_t $.
    		Following the definition in~\eqref{eq:Rdef_v2} for $R$, we analyze the expected regret $\E [R]$ in the new semi-bandit feedback setting and CDN loss sequence $ L_{1:T}$ given in Algorithm~\ref{alg:sequence2}. Let $\mc Q_0$ and $\mc Q_{\mc I}$ follow previous definition in~\eqref{eq:Qdef1} and~\eqref{eq:Qdef2}. Let $\mc F$ be the $\sigma $-algebra generated by the observations $Z_{1:T} $, where $Z_t = L_t \circ A_t$. The total variation  $d_{\mathrm{TV}}^{\mc F}$ is defined as in~\eqref{eq:defdev_v3} with respect to the $\sigma $-algebra $\mc F$.
			For $Y_{0:T}$ defined in~\eqref{eq:defyseq} of~\S\ref{subsec:lowerbound-semi} and the action sequence $A_{1:T}$, we define	\begin{align*}
			R' \triangleq 	\sum_{t=1}^T  \langle  A_t,\tilde{L}_t\rangle+  \lambda \sum_{t=1}^T   d( A_t,\tilde A_{t-1}) - \min_{A\in \mc A} \sum_{t=1}^T   \langle A
   , \tilde{L}_t  \rangle.
		\end{align*}
   \begin{remark}\label{rm:algebra2}
	    Note that $A_t$ is a deterministic function of its past observations $Z_{1:t-1}$, thus the $\sigma$-algebra generated by $A_{1:T}$ is a subset of $\mc F.$
	\end{remark}	

		\begin{lemma}\label{cor:inequa2}
			It holds that
			$$\frac{1}{\binom{K}{I}}  \sum_{\mathcal I \subset [K]} d^{\mathcal F}_{\mathrm{TV}} (\mathcal Q_0, \mathcal Q_{\mathcal I}) \leq  \frac{\epsilon }{\sigma \sqrt{2K}} \sqrt{I(\log_2 T+1) \E_{\mathcal Q_0}[M]}.$$
		\end{lemma}
  \begin{IEEEproof}
      	 By Lemma~\ref{lm:lowerboundsemi},  Pinsker’s inequality~\cite[Lemma 11.6.1]{cover1999elements} and $w(\rho)\leq \log_2 T+1$, we have $$	d^{\mathcal F'}_{\mathrm{TV}} (\mathcal Q_0, \mathcal Q_{\mathcal I})  \leq \frac{\epsilon}{2\sigma} \sqrt{(\log_2 T+1) \E_{\mathcal Q_0}[M_{\mathcal I}]}  ,$$ where $\mc F'$ is the $\sigma$-algebra generated by $Y_{0:T}$ (defined in~\eqref{eq:defyseq} of~\S\ref{subsec:lowerbound-semi}).  Since $Z_{1:T}$ is a function of $Y_{0:T}$, we have $\mathcal F\subset \mathcal F'$ which implies $d^{\mathcal F}_{\mathrm{TV}} (\mathcal Q_0, \mathcal Q_{\mathcal I})\leq\frac{\epsilon}{2\sigma} \sqrt{(\log_2 T+1)  \E_{\mathcal Q_0}[M_{\mathcal I}]}.$ Then using the concavity of the squared root function and by $ \sum_{\mathcal I \subset [K]} M_{\mathcal I} = 2\binom{K-1}{I-1} M$, it holds that
			\begin{align*}
				&\frac{1}{\binom{K}{I}}  \sum_{\mathcal I \subset [K]}	d^{\mathcal F}_{\mathrm{TV}} (\mathcal Q_0, \mathcal Q_{\mathcal I}) \nonumber\\*
    &\leq \frac{\epsilon}{2\sigma} \sqrt{ \log_2 T+1 }  \sum_{\mathcal I \subset [K]} \frac{1}{\binom{K}{I}} \sqrt{ \E_{\mathcal Q_0}[M_{\mathcal I}]} \\
				&\leq \frac{\epsilon}{2\sigma} \sqrt{\log_2 T+1  } \sqrt{ \E_{\mathcal Q_0}\Bigg[ \frac{1}{\binom{K}{I}}  \sum_{\mathcal I \subset [K]}M_{\mathcal I}\Bigg]}\\
				&= \frac{\epsilon}{\sigma\sqrt{2K}} \sqrt{I (\log_2 T+1) \E_{\mathcal Q_0}[  M] }.
			\end{align*}
  \end{IEEEproof}
		\begin{lemma}\label{lm:RRprime2}
			Assume that $ T\geq  \max \{\frac{\lambda K}{I^2},\frac{I}{6}\} $. Then $ 	\E[R]\geq 	\E[R'] -\frac{\epsilon TI}{6} $.
		\end{lemma}
		\begin{IEEEproof}
			We consider the event $ B=\{\forall\, t: L_t=\tilde{L}_t \} $, and first show that $ P(B)\geq 5/6 $. For $ \delta = \frac{I}{T} \leq \frac{1}{6} $, by~\cite[Lemma 1]{dekel2014bandits} we have that with probability at least $ \frac{5}{6} $,  for all $ t\in [T] $ and $i\in [K]$,
			\begin{align*}
				|W^i_t|\leq\sigma \sqrt{2d(\rho)\log_2 \frac{TI}{\delta}} \leq 3\sigma  \log_2 T,
			\end{align*}
			where the last inequality is due to $d(\rho)\leq \log_2 T+1$ by~\cite[Lemma 2]{dekel2014bandits}. Thus, setting $ \sigma =\frac{1}{9\log T} $ we obtain that 
			\begin{align*}
				P\Big( \forall\, t\in [T], i\in [K] , \frac{1}{2} +W^i_t\in [\frac{1}{6},\frac{5}{6}]\Big)\geq \frac{5}{6}.
			\end{align*}
			For $ T\geq\max(\frac{\lambda K}{I^2},6) $, we have $ \epsilon < \frac{1}{6} $ and thus $ \tilde{L}_t(x)\in [0,1] $ for all $ x\in [K] $ whenever $ \frac{1}{2}+W_t \in [\frac{1}{6},\frac{5}{6}] $. This implies that $ P(B)\geq \frac{5}{6}. $
			
			If $ B $ occurs, $ R=R' $; otherwise, $\lambda M\leq R\leq R'\leq \lambda M+\epsilon TI $, so that $ R'-R\leq \epsilon TI $. Therefore,
			\begin{align*}
				\E[R']-\E[R] = \E[R'-R| \lnot B]\cdot P(\lnot B)\leq \frac{\epsilon TI}{6}.
			\end{align*}
		\end{IEEEproof}
		
		\begin{lemma}\label{lm:Rprime2}
  It holds that
			\begin{align*}
				\E[R'] \geq \epsilon TI \Big(1-\frac{I}{K}\Big) - \frac{\epsilon  TI}{\binom{K}{I}} \sum_{\mathcal I\in \mc A} d^{\mathcal F}_{\mathrm{TV}} (\mathcal Q_0,\mathcal Q_{\mathcal I}) + \lambda\E[M]
			\end{align*}
		\end{lemma}
		\begin{IEEEproof}
			For any $ i\in [K]$, let $ T_{ i} $ denote the number of times the player picks arm $i$ when  the time horizon is $T$.
			So we can write $ R'  = \epsilon \Big(TI -\sum_{i\in [K]} \chi_i  T_i\Big) +\lambda M $, where $\chi_i$ is the $i$-th component of $\chi$. We also use $\mc I_i$ is the $i$-th component of $\mc I\in \mc A$. Thus
			\begin{align*}
				\E[R'] &= \frac{1}{\binom{K}{I}} \sum_{\mc I\in \mc A} \E\Big[   \epsilon \Big(TI -\sum_{i\in [K]}\mc I_i T_i\Big) +\lambda M  \Big | \chi=\mc I \Big]\\
				&= \epsilon TI - \frac{\epsilon}{\binom{K}{I}} \sum_{\mc I\in \mc A} \sum_{i\in [K]} \mc I_i\E_{Q_{\mc I}}\big[T_i\big] +\lambda\E[M].
			\end{align*}
			For all $ i\in [K]$ and $ t\in [T] $, the event $\{ A_{t,i}=1 \} $ belongs to the $ \sigma $-field $ \mathcal F $ by Remark~\ref{rm:algebra2}, so 
			\begin{align*}
				\mathcal Q_{\mathcal I}(A_{t,i}=1 ) - \mathcal Q_0(A_{t,i}=1) \leq d^{\mathcal F}_{\mathrm{TV}} (\mathcal Q_0,\mathcal Q_{\mathcal I}).
			\end{align*}
			Summing over $ t\in [T] $ yields
			\begin{align*}
				\E_{\mathcal Q_{\mathcal I}} \big[T_i\big] - 	\E_{\mathcal Q_{0}} \big[T_i\big] \leq Td^{\mathcal F}_{\mathrm{TV}} (\mathcal Q_0,\mathcal Q_{\mathcal I}).
			\end{align*}
   Summing over $ i\in [K] $ such that $\mc I_i=1$ yields
			\begin{align*}
				\sum_{i\in [K]}	\mc I_i\Big(	\E_{\mathcal Q_{\mathcal I}} \big[T_i\big] - 	\E_{\mathcal Q_{0}} \big[T_i\big] \Big)\leq TId^{\mathcal F}_{\mathrm{TV}} (\mathcal Q_0,\mathcal Q_{\mathcal I}).
			\end{align*}
			 Summing over $ \mc I\in \mc A $ yields
			\begin{align*}
				\sum_{\mathcal I\in \mc A}\sum_{i\in [K]} \mc I_i\Big(	\E_{\mathcal Q_{\mathcal I}} \big[T_i\big] - 	\E_{\mathcal Q_{0}} \big[T_i\big]\Big)\leq TI	\sum_{\mathcal I\in \mc A} d^{\mathcal F}_{\mathrm{TV}} (\mathcal Q_0,\mathcal Q_{\mathcal I}).
			\end{align*}
			Thus
			\begin{align*}
				&\sum_{\mathcal I\in \mc A}\sum_{i\in [K]} \mc I_i\E_{\mathcal Q_{\mathcal I}}  \big[T_i\big] \nonumber\\*
    &\leq 	\sum_{\mathcal I\in \mc A}\sum_{i\in [K]}	\mc I_i\E_{\mathcal Q_{0}}  \big[T_i\big]  + TI	\sum_{\mathcal I\in \mc A} d^{\mathcal F}_{\mathrm{TV}} (\mathcal Q_0,\mathcal Q_{\mathcal I}) \\
				& =\binom{K-1}{I-1} TI+TI	\sum_{\mathcal I\in \mc A} d^{\mathcal F}_{\mathrm{TV}} (\mathcal Q_0,\mathcal Q_{\mathcal I})
			\end{align*}
			Therefore,
			\begin{align*}
				\E[R'] &\geq \epsilon TI - \frac{\epsilon}{\binom{K}{I}}\left( \binom{K-1}{I-1} TI+TI	\sum_{\mathcal I\in \mc A} d^{\mathcal F}_{\mathrm{TV}} (\mathcal Q_0,\mathcal Q_{\mathcal I})\right)  \nonumber\\*
    & \qquad+ \lambda\E[M]\\
				&=\epsilon TI \Big(1-\frac{I}{K}\Big) - \frac{\epsilon  TI}{\binom{K}{I}} \sum_{\mathcal I\in \mc A} d^{\mathcal F}_{\mathrm{TV}} (\mathcal Q_0,\mathcal Q_{\mathcal I}) + \lambda\E[M]
			\end{align*}
			
		\end{IEEEproof}
		\begin{IEEEproof}[Proof of Lemma~\ref{lm:lowerbound2}]
			We first prove the theorem for deterministic players that make no more than $S_0= \epsilon T I/\lambda$ switches. For algorithms with this property, we have 
			\begin{align*}
				\mathcal Q_0\Big(M>  \frac{\epsilon T I}{\lambda}\Big)=	\mathcal Q_{\mathcal I}\Big(M>  \frac{\epsilon T I}{\lambda}\Big) =0.
			\end{align*}
   By Remark~\ref{rm:algebra2}, the event $\{M\geq m\}\in \mc F$ which implies 
			\begin{align*}
				\E_{\mathcal Q_0}[M] - \E_{\mathcal Q_{\mathcal I}}[M]  &= \sum_{m=1}^{\epsilon TI/\lambda} \big(\mathcal Q_0 (M\geq m)- \mathcal Q_{\mathcal I} (M\geq m) \big) \\
				&\leq \frac{\epsilon TI}{\lambda} d^{\mathcal F}_{\mathrm{TV}} (\mathcal Q_0, \mathcal Q_{\mathcal I}).
			\end{align*}
			Then 
			\begin{align*}
				\E_{\mathcal Q_0}[M] - \E[M]   &= \frac{1}{\binom{K}{I}} \sum_{\mathcal I \in \mc A}
				\Big( \E_{\mathcal Q_0}[M]- \E_{\mathcal Q_{\mathcal I}}[M] \Big)\\
				&\leq \frac{ \epsilon TI}{\lambda\binom{K}{I}} \sum_{\mathcal I \in \mc A}d^{\mathcal F}_{\mathrm{TV}} (\mathcal Q_0, \mathcal Q_{\mathcal I})
			\end{align*}
			Combining this with Lemma~\ref{lm:RRprime2} and Lemma~\ref{lm:Rprime2}, we obtain
			\begin{align*}
				\E[R]&\geq \epsilon \Big(1-\frac{I}{K}-\frac{1}{6}\Big) -  \frac{2\epsilon  TI}{\binom{K}{I}} \sum_{\mathcal I\in \mc A} d^{\mathcal F}_{\mathrm{TV}} (\mathcal Q_0,\mathcal Q_{\mathcal I})  \nonumber\\*
    &\qquad+ \lambda \E_{\mathcal Q_0} [M].
			\end{align*}
			By Corollary~\ref{cor:inequa2}, and $ \log_2 T+1\leq 2\log_2 T $,
			\begin{align*}
				\frac{1}{\binom{K}{I}}  \sum_{\mathcal I \subset [K]} d^{\mathcal F}_{\mathrm{TV}} (\mathcal Q_0, \mathcal Q_{\mathcal I}) 
				&\leq   \frac{\epsilon  }{\sigma \sqrt{K}} \sqrt{I(\log_2 T) \E_{\mathcal Q_0}[M]}.
			\end{align*}
			Using the notation $ m= \sqrt{\E_{\mathcal Q_0}[M]} $ and when $ K\geq 3 I $,
			\begin{align*}
				\E[R] \geq \frac{\epsilon TI}{2} - \frac{2\epsilon^2  TI^{3/2}}{\sigma \sqrt{K}} \sqrt{\log_2 T }m +\lambda m^2,
			\end{align*}
			where the right hand side is minimized when $ m = \frac{\epsilon^2 TI^{3/2} \sqrt{\log_2 T}}{\lambda\sigma \sqrt{K}} $. Thus the right hand side is lower bounded by $   \frac{\epsilon TI}{2} - \frac{\epsilon^4 T^2I^3 \log_2 T}{\lambda\sigma^2 K}$. Using our choice of $ \sigma =\frac{1}{9 \log_2 T} $ and $ \epsilon = \frac{(\lambda K)^{\frac{1}{3}}I^{-\frac{2}{3}}T^{-\frac{1}{3}}}{9\log_2 T} $, gives
			\begin{align}\label{eq:erineq}
				\E[R]\geq \frac{(\lambda K I)^{\frac{1}{3}} T^{\frac{2}{3}}}{30\log_2 T}.
			\end{align}

   	For any general algorithm that has an arbitrary number of switches, we can turn it to a new algorithm that makes at most $S_0$ switches by halting the algorithm once it makes $S_0$ switches and repeating the last action in the remaining rounds. The regret $R^*$ of the new algorithm equals $R$ when $M\leq S_0$ and when $M> S_0$,
			\begin{align*}
				R^* \leq R+\epsilon TI \leq 2R,
			\end{align*}
			since $R\geq \lambda S_0$. Thus $\E[R^*]\leq 2\E[R]$. Since $\E[R^*]$ is lower bounded by the right-hand side of \eqref{eq:erineq}, this implies the claimed lower bound on the expected regret of any deterministic player.
		\end{IEEEproof}
\bibliographystyle{IEEEtran}
		\bibliography{paper}

\begin{IEEEbiographynophoto}
    {Yanyan Dong} received her B.S.\ degree from Jilin University in 2017, and Ph.D.\ degree from The Chinese University of Hong Kong, Shenzhen in 2022. She was a Research Fellow at the National University of Singapore from Dec.~2022 to Dec.~2023. Her research interests include information theory, coding theory, network coding, and machine learning.
\end{IEEEbiographynophoto}

\begin{IEEEbiographynophoto}{Vincent Y.\ F.\ Tan} (Senior Member, IEEE)   was born in Singapore in 1981. He received the B.A.\ and M.Eng.\ degrees in electrical and information science from Cambridge University in 2005, and the Ph.D.\ degree in electrical engineering and computer science (EECS) from the Massachusetts Institute of Technology (MIT) in 2011. He is currently a Professor with the Department of Mathematics and the Department of Electrical and Computer Engineering (ECE), National University of Singapore (NUS). His research interests include information theory, machine learning, and statistical signal processing.

Dr.\ Tan is an elected member of the IEEE Information Theory Society Board of Governors. He was an IEEE Information Theory Society Distinguished Lecturer from 2018 to 2019. He received the MIT EECS Jin-Au Kong Outstanding Doctoral Thesis Prize in 2011, the NUS Young Investigator Award in 2014, the Singapore National Research Foundation (NRF) Fellowship (Class of 2018), and the NUS Young Researcher Award in 2019.  He is currently serving as a Senior Area Editor for the {\em IEEE Transactions on Signal Processing} and as an Associate Editor in Machine Learning and Statistics for the {\em IEEE Transactions on Information Theory}. He also regularly serves as an Area Chair of prominent machine learning conferences such as the {\em   International Conference on Learning Representations} (ICLR) and the {\em Conference on Neural Information Processing Systems} (NeurIPS).
\end{IEEEbiographynophoto}

	\end{document}